\documentclass[final]{cvpr}
\usepackage{times}
\usepackage{epsfig}
\usepackage{graphicx}
\usepackage{amsmath}
\usepackage{amssymb}
\usepackage{comment}
\usepackage[pagebackref=true,breaklinks=true,colorlinks,bookmarks=false]{hyperref}
\usepackage{array,multirow,textcomp}
\usepackage[skip=0.5\baselineskip]{caption}
\usepackage{xcolor}

\def\cvprPaperID{8339} 
\def\confYear{CVPR 2021}
\begin{document}
\title{Efficient Conditional GAN Transfer with Knowledge Propagation across Classes}

\author{Mohamad Shahbazi$^1$, Zhiwu Huang$^1$, Danda Pani Paudel$^1$, Ajad Chhatkuli$^1$, Luc Van Gool$^{1, 2}$\\
$^1$Computer Vision Lab, ETH Zürich, Switzerland \quad $^2$PSI, KU Leuven, Belgium\\
{\tt\small \{mshahbazi, zhiwu.huang, paudel, ajad.chhatkuli, vangool\}@vision.ee.ethz.ch}
}
\maketitle

\begin{abstract}
Generative adversarial networks (GANs) have shown impressive results in both unconditional and conditional image generation. In recent literature, it is shown that pre-trained GANs, on a different dataset, can be transferred to improve the image generation from a small target data. The same, however, has not been well-studied in the case of conditional GANs (cGANs), which provides new opportunities for knowledge transfer compared to unconditional setup. In particular, the new classes may borrow knowledge from the related old classes, or share knowledge among themselves to improve the training. This motivates us to study the problem of efficient conditional GAN transfer with knowledge propagation across classes. To address this problem, we introduce a new GAN transfer method to explicitly propagate the knowledge from the old classes to the new classes. The key idea is to enforce the popularly used conditional batch normalization (BN) to learn the class-specific information of the new classes from that of the old classes, with implicit knowledge sharing among the new ones. This allows for an efficient knowledge propagation from the old classes to the new ones, with the BN parameters increasing linearly with the number of new classes. The extensive evaluation demonstrates the clear superiority of the proposed method over state-of-the-art competitors for efficient conditional GAN transfer tasks. The code is available at:
\url{https://github.com/mshahbazi72/cGANTransfer}
\end{abstract}

\section{Introduction}
Generative adversarial networks (GANs)~\cite{goodfellow2014gans, arjovsky2017wgan} are the most common models used for image and video generation, showing very promising results both~\cite{liu2020generative} in unconditional~\cite{karras2017progressive,wu2019sliced,karras2019style} and conditional~\cite{brock2018large,park2019semantic,clark2019adversarial} setups.

Learning from limited data is a well-studied problem in the discriminative setup, where the concept of knowledge transfer~\cite{pan2009survey} between two different but related tasks~\cite{zamir2018taskonomy} or domains~\cite{ganin2015unsupervised} is ubiquitous. 
In contrast, the literature on transfer learning for generative adversarial models is fairly limited. One may find this unexpected, since many popular knowledge transfer methods in discriminative setup, in turn, use generative schemes~\cite{liu2016coupled, tzeng2017adversarial}. However, the limited literature is less surprising when the complexity of adversarial training and the mode collapse are taken into account.

A notable work by \emph{Wang et al.}~\cite{wang2018transferring}, first addressed the problem of training GANs on limited data using a careful fine-tuning (FT) strategy. Following works~\cite{mo2020freeze,wang2020minegan,zhao2020leveraging} are the variants of~\cite{wang2018transferring} that focus on better fine-tuning strategies. 
On the contrary, \emph{Noguchi \& Harada}~\cite{noguchi2019image} proposed the batch statistics adaptation (BSA) technique, by learning only the batch normalization parameters on a small target dataset. 
As most of the previous works~\cite{wang2018transferring,wang2020minegan,noguchi2019image} primarily focus on the case of unconditional GANs, we investigate in a different direction of conditional GANs (cGANs). In particular, we are interested in producing new classes given a pre-trained class-conditional cGAN. cGANs are strikingly interesting due to their capability of handling a large number of classes with a single network. For example, \emph{BigGAN}~\cite{brock2018large} can generate images from all 1K classes of ImageNet~\cite{deng2009imagenet}. In fact, \emph{BigGAN} is exploited as the pre-trained network even by the unconditional methods~\cite{wang2020minegan,noguchi2019image}. We refrain from fine-tuning whenever possible, as we believe that new classes can be introduced within such powerful cGANs. Moreover, some powerful pre-trained cGANs, 
can potentially be used to add new classes in the lifelong learning~\cite{li2017learning, noguchi2019image,parisi2019continual,cong2020ganmemory} fashion, which however is beyond the focus of our paper.

In this work, we study how new classes with a limited amount of samples can be added to pre-trained cGANs using knowledge transfer across classes. To do so, inspired by BSA~\cite{noguchi2019image}, we aim at learning only the batch normalization (BN) parameters that generally encode the class-specific information. Our key idea, different from~\cite{noguchi2019image}, relies on the assumption that the knowledge between old and new classes can also be transferred by searching for the similarity between them. Our experimental setup, however, does not allow us to access the old data used for the pre-trained model. Therefore, the similarity is searched in the conditional space of the BN parameters, during the training of cGANs. In this process, we learn the similarity scores \emph{explicitly between old and new classes, and implicitly between new ones}.
The learned old-to-new similarity scores are then used to derive the batch statistics of new classes from that of old ones. 

It is well-established in~\cite{bengio2009curriculum,hinton2015distilling,vinyals2016matching,li2017learning,sung2018learning,guo2017active, data2018interpolating} and many other works, that learning algorithms can greatly benefit from the shared knowledge between classes. Often, such similarity is either known or discovered when all the classes are accessible. In the context of domain generalization (or in some special case of adaptation), the source data is similarly inaccessible partially or completely~\cite{panareda2017open,zhou2020domain,liu2020open}. However, the latter assumes that the new classes are either the same or largely overlap with the old ones. Note that in our case, \emph{new classes do not even overlap with the inaccessible old ones}. In addition, almost all aforementioned works seek similarity in the feature space with an exception of~\cite{data2018interpolating}. However, ~\cite{data2018interpolating} is primarily  designed  to serve the
discriminative models.
Our generative case, on the other hand, hinders us to access the feature space. We, therefore, rely on the conditional space of cGANs to establish the sought class similarities. Up to our knowledge, we learn the inter-class similarities in the conditional space of the generative models, for the first time.

In summary, we utilize cross-class knowledge while introducing new classes in cGANs. While doing so, active searching of similarity scores between new and old classes with implicit knowledge sharing among new ones is suggested. In this context, we propose a novel method for finding the similarity between new and old classes without requiring access to the old data. The proposed method is particularly suitable when transferring knowledge from pre-trained cGANs. 

In summary, the key contributions of our work are as follows:
\begin{itemize}
    \item We study the new problem of efficient GAN transfer to new classes with explicit inter-class knowledge propagation in pre-trained cGANs. \vspace{-0.2cm}
    \item A novel method for learning similarity between old and new classes and knowledge sharing within the new classes is proposed using the batch normalization statistics of the old classes, in the conditional space of generative models. 
    \vspace{-0.2cm}
    \item Our experiments on three benchmark datasets demonstrate the superiority of our method both in terms of generated image quality and the convergence speed. 
\end{itemize}
\section{Related Work}
\noindent\textbf{Class-conditional GANs.}
Different architectures and loss functions have been proposed for conditioning GANs on class labels~\cite{mirza2014cgan, miyato2018cganproj, kavalerov2019cgan}. The current state-of-the-art methods for class conditioning commonly employ cGAN with projection discriminator~\cite{miyato2018cganproj, miyato2018spectralnorm}. In the generator, conditional batch normalization~\cite{devries2017modulating} with class-specific scale and shift parameters are applied to each layer of the generator. The discriminator, on the other hand, is conditioned on the class labels by computing the dot product of the last feature layer and the learnable embedding of the desired class. The performance of the conditional GANs was further improved by adding self-attention layers to the generator and the discriminator~\cite{zhang2019sagan}. BigGAN~\cite{brock2018large} was able to reach state-of-the-art performance on image generation from ImageNet, mainly by using a bigger batch size (2048) and some architectural improvements such as a hierarchical latent variable. The conditioning, however, still happens through the class-conditional batch normalization in the generator and the projection layer in the discriminator.

\noindent\textbf{Transfer Learning in GANs.}
Iterative image generation approaches, such as DGN-AM~\cite{nguyen2016DGNAM} and PPGN~\cite{nguyen2017PPGN}, could be considered as early attempts on transfer learning in image generation by generating images via maximizing the activation of the neurons of a pre-trained classifier. TransferGAN~\cite{wang2018transferring} is one of the earliest studies addressing transfer learning in GANs. The authors showed that, by simply fine-tuning a pre-trained network on the target dataset, they can outperform training from the scratch in terms of image quality and convergence time. However, naive fine-tuning on small data still suffers from mode collapse and training instability. Another method~\cite{zhao2020leveraging} proposes transferring the low-level layers of the generator and the discriminator from the pre-trained network, while learning the high-level layers from scratch for the target data. In a recent study~\cite{mo2020freeze}, it is shown that simply freezing the low-level filters of the discriminator is more effective than previous fine-tuning approaches. BSA~\cite{noguchi2019image}, on the other hand, instead of looking for ways of fine-tuning the network, proposed freezing the weights of the pre-trained generator except the batch normalization parameters. For the target data, new BN parameters are learned without fine-tuning the generator. This allows BSA to add new classes without disturbing the old ones. 
MineGAN~\cite{wang2020minegan} learns a small fully-connected miner network at the input of a frozen pre-trained GAN. The miner learns to shift the prior input distribution to the most suitable one for the target data. After training the miner, MineGAN further fine-tunes both the generator and the miner as the final model. MineGAN is designed to transfer knowledge to a single-class target.

\begin{figure*}[t]
\begin{center}
   \includegraphics[width=1\linewidth]{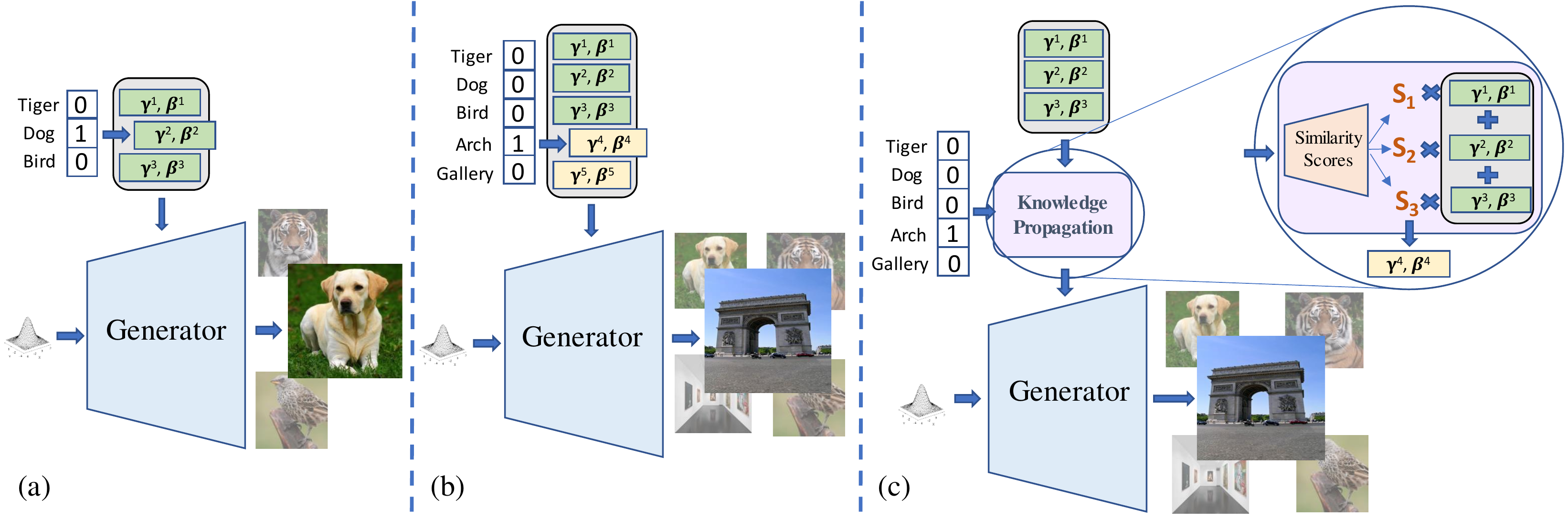}
\end{center}
   \caption{(a): A generator trained on a multi-class dataset. (b): The generator, extended by new classes without taking the old classes into account. (c): The conceptual diagram of our proposed method for conditional GAN transfer with knowledge propagation across classes. Here $\gamma^i$ and $\beta^i$ ($1 \leq i \leq 5$) represent the class-specific parameters.}
\label{fig:conceptual}
\end{figure*}

\noindent\textbf{Inter-class Knowledge Transfer.}
A lot of works have been emerging on extending models trained on previous examples/images to perform favorably on new data, and here is where knowledge transfer becomes essential. For instance, \cite{vinyals2016matching} uses memory and attention modules to transfer labeled data knowledge to a new class example. A more relevant work \cite{guo2017active} uses the class similarity between source classes and a target class to improve the classification performance. \cite{sung2018learning} is a few-shot meta-learning method that uses the so-called relation score of a new class example with previous examples in order to classify the new example. In a more related work~\cite{data2018interpolating}, BN layers have been used for knowledge transfer from old classes, mainly targeting binary classification with a brief empirical study on style transfer. The knowledge transfer framework of~\cite{data2018interpolating} separately performs encoding, pre-selection, and combination of old classes. In contrast, our method jointly optimizes the prior knowledge and the transfer process, leading to an efficient learning paradigm for the generative case.

\section{Problem Definition}\label{sec:probdef}
The task of class-conditional GAN transfer aims at approximating the target data distribution by transferring knowledge from the source multi-class data to the target data using pre-trained GAN models. As shown in Fig.~\ref{fig:conceptual} (a), the pre-trained GAN model consists of one generator $G$ and one discriminator $D$, jointly trained on the source multi-class dataset $\mathcal{X}=\{X^1, X^2, \dots,X^N\}$, in which $X^y$ is a set of real images in category $y\in\{1, 2, \dots, N\}$ with the underlying distribution $P_y(x)\in\{P_1(x), P_2(x), \dots, P_N(x)\}$. In the conditional setup, given a random noise vector $z$ (usually sampled from $\mathcal{N}(0,I)$) and a class label $y$ as inputs, $G$ is trained in an adversarial game with $D$, to generate an image $x\sim P_y(x)$:
\begin{equation}
\begin{split}
    &x = G(z, y); \\
    &z\sim \mathcal{N}(0,I), \; y\in \{1, \dots,N\} \quad
    \text{s.t.} \,\, x \sim P_y(x).
\end{split}
\end{equation}

 In the state-of-the-art conditional GANs (\eg, \cite{miyato2018cganproj, zhang2019sagan, brock2018large}), $G$ is commonly conditioned on the class labels using conditional batch normalization. Specifically, the layer-wise output $f_l$, 
of layer $l \in \{1, \dots,L\}$,  is normalized and modulated by the class-specific scale $\gamma^y_l \in \{\gamma^1_l, \dots, \gamma^N_l\}$ and shift $\beta^y_l \in \{\beta^1_l, \dots, \beta^N_l\}$ as:
\begin{equation}\label{eq2}
\begin{split}
    f'_l = \gamma^y_l \frac{f_l-\mu_l}{\sigma_l} + \beta^y_l,
\end{split}
\end{equation}
where $\mu_l$ and $\sigma_l$ represent the batch mean and variance for the $l$-th layer, and $f'_l$ is the normalized output of the layer . Thus, the class-specific information is parameterized by the corresponding scales and shifts.

{Transfer learning in conditional GANs} can be defined as exploiting pre-trained GAN models to adapt the generator/discriminator to a new multi-class target data ${\mathcal{X}'=\{X^{N+1}, X^{N+2}, \dots, X^{N+M}\}}$ with $M$ new categories $\{y'\},\ y'\in\{N+1,\hdots, N+M\}$. Mathematically, this task can be defined as the following learning problem:
\begin{equation}
\label{eq:finetuning}
\begin{split}
    &x' = G'(z, y'); \\
    &z\sim \mathcal{N}(0,I), \; y'\in \{N+1, ...,N+M\}\\
    & \text{s.t.}\,\, x' \sim P_{y'}(x),\ \ \text{given } G(z, y) 
,\end{split}
\end{equation}
where $G'$ is the new generator learned for the new categories $\{y'\}$. It is often desirable that  $G'$ can also generate the previous classes, as shown in Fig.~\ref{fig:conceptual} (b). The problem of learning such a generator which is capable of generating images for both $\{y\}$ and $\{y'\}$ can be formulated as:
\begin{equation}
\label{eq:classext}
\begin{split}
    &x = G'(z, y_f); \\
    &z\sim \mathcal{N}(0,I), \; y_f\in \{1, ...,N+M\} \\
    &\text{s.t.}\,\, x \sim P_{y_f}(x),\ \ \text{given }  G(z, y), 
\end{split}
\end{equation}
where we denote the final category set as $\{y_f\}= \{y\} \cup \{y'\}$ and the extended generator as $G'$.  

Transfer learning  in the context of GANs is usually approached by a careful fine-tuning of a pre-trained model~\cite{wang2018transferring, wang2020minegan, mo2020freeze, zhao2020leveraging}, with or without modifying the architecture. These architecture modifications include adding either new layers or new batch normalization parameters, to additionally learn the new knowledge of the target data. Such approaches overlook the shared similarities among the old and new classes, resulting in an inefficient knowledge propagation.
This motivates us to aim for a more efficient conditional GAN transfer. In particular, we seek for an explicit knowledge propagation from old (\ie, source) classes to the new (\ie, target) ones, along with the possibility of knowledge sharing among target classes. 
Fig.~\ref{fig:conceptual} (c) conceptually illustrates the problem addressed in this paper.

\section{Knowledge Transfer Across Classes}\label{sec:method}
We first provide an overview of the proposed method, followed by the details. Our premise is built upon a known observation in BigGAN~\cite{brock2018large}, where interpolating between different classes produces \emph{visually meaningful} intermediate images that do not exist in the training data. This implies that the learned class-specific parameters could lie on a smooth manifold, on which new classes also reside. 
Consequently, it begs the question: can the similarities between the source and the target classes--as well as those within the target--be exploited to learn the target representations? This question leads us to learn the parameters of the new classes, while being dependent upon the parameters (representing the knowledge) of old classes. More precisely, we propose to obtain the representation of each target class by learning a suitable linear combination over the representations of the source classes -- which we call \emph{knowledge propagation}. In addition, we propose a mechanism to enable \emph{knowledge sharing} within target classes, by optimizing the source knowledge in favor of the multiple target classes\footnote{Although the focus of our work is on the multi-class target data, our method can be generalized well to the single-class target as well (See the supplementary material, Sec. F).}.
We address the exact problem of~\eqref{eq:classext}, where the model aims to generate both new and old classes. Such consideration is often ignored in the literature, with an exception of BSA~\cite{noguchi2019image}. 
The details of the proposed knowledge propagation and sharing strategies are presented in Sec.~\ref{subSec:propagation}  and Sec.~\ref{subSec:sharing}, respectively.  

\begin{figure}[t]
\begin{center}
   \includegraphics[width=1\linewidth]{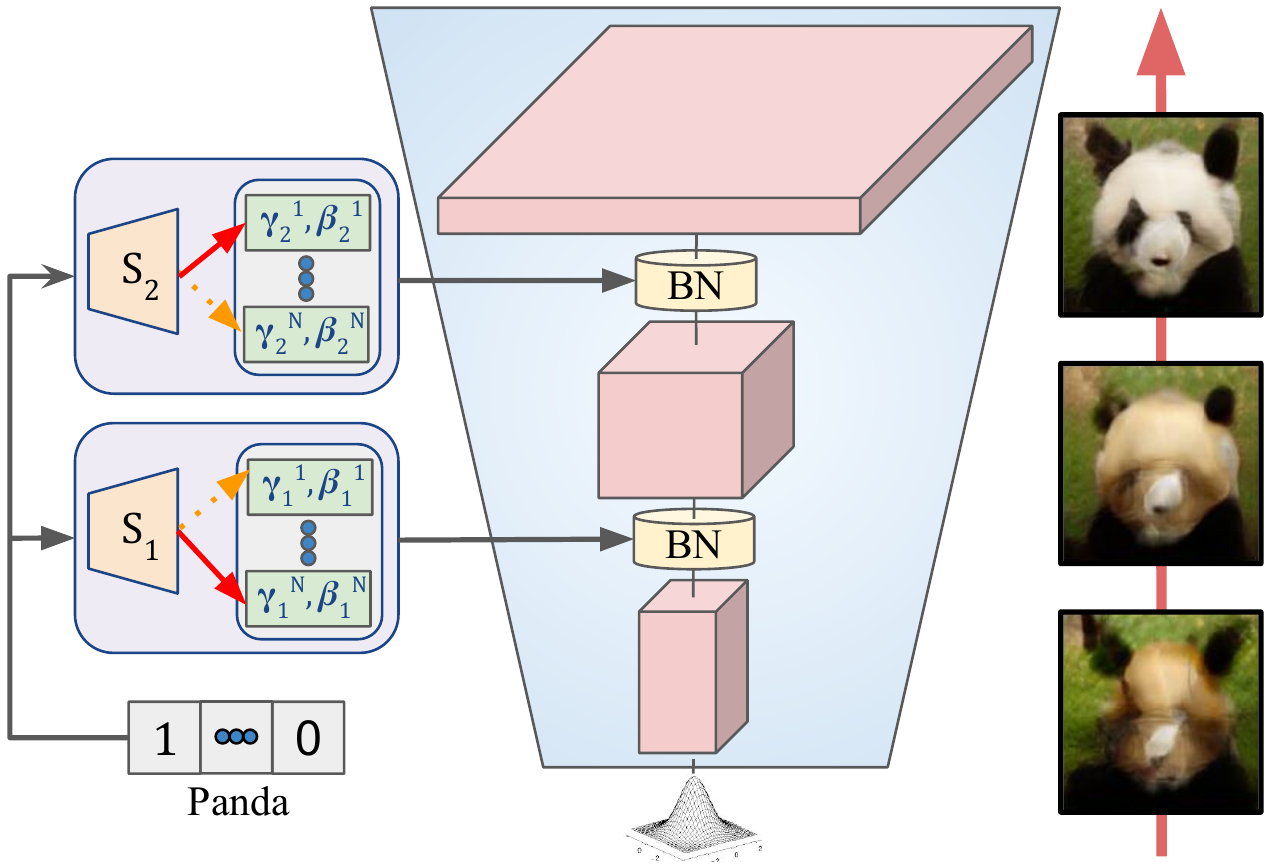}
\end{center}
   \caption{Intuitive visualization of how different layers might borrow information from different classes based on the hierarchy of features such as shape and color (for a better visualization, only two layers are illustrated). Pictures in the figure are obtained from our experiment on AnimalFace dataset~\cite{Zhangzhang2011animal} (refer to Fig.~\ref{fig:spectrum} for more visualizations).}
\label{fig:layerwise}
\end{figure}

\subsection{Knowledge Propagation}\label{subSec:propagation}
Following the class-conditioning paradigm discussed in Sec.~\ref{sec:probdef}, we embed the class-specific representations in the batch normalization (BN) layers as the scale and shift parameters. Accordingly, knowledge propagation is performed in BN layers. To obtain the BN parameters ($\gamma_l$ and $\beta_l$ in~\eqref{eq2}) of each new class, our model linearly combines the BN parameters of old classes $\{y\}$ using layer-wise similarity scores, learned during the training step:
\begin{equation}\label{eq:lincomb}
\begin{split}
    &\gamma^{y'}_l = [\gamma^1_l, \gamma^2_l, \dots, \gamma^N_l]  {S^{y'}_{\gamma_l}}, \\
    &\beta^{y'}_l = [\beta^1_l, \beta^2_l, \dots, \beta^N_l] {S^{y'}_{\beta_l}}, \\
    &y' \in \{N+1, \dots, N+M\}, \;\; l \in \{1, \dots, L\}.
\end{split}
\end{equation}
Here, ${S^{y'}_{\gamma_l}} = [s_{{\gamma_l}}^{(y',1)},\dots,s_{{\gamma_l}}^{(y',N)}]^\top \in \mathbb{R}^N$ and similarly ${{S^{y'}_{\beta_l}}\in \mathbb{R}^N}$ are vectors of learned scores for the class $y'$ in layer $l$. Two things are to be noted here. First, we learn the similarity scores for scale and shift parameters separately. This is because, the new classes could be similar to some of the old classes in terms of their distribution mean, while being similar to another set of classes in terms of intra-class variance. 
Secondly, we also learn a different set of scores per layer, since different layers of the network do not necessarily benefit from the same set of old classes. It is well-known that different layers of neural networks represent different levels of feature representation. As intuitively shown in Fig.~\ref{fig:layerwise}, some layers could be responsible for the general shape, some for the color and the texture, and some others for finer details. Such hierarchy of features is also the main motivation of StyleGAN~\cite{karras2019style} for using layer-specific styles.

Based on the empirical observations, 
we propagate class-specific knowledge only in the generator and simply fine-tune the discriminator. A similar knowledge propagation in the discriminator leads to performance degradation in our experiments. One possible reason for such degradation is that the knowledge propagation speeds up the over-fitting problem of the discriminator on small datasets~\cite{zhao2020differentiable}.

\subsection{Knowledge Sharing}\label{subSec:sharing}
In addition to knowledge propagation, we propose a mechanism for knowledge sharing among target classes. 
Consider transferring a conditional GAN, pre-trained on ImageNet classes, to the AnimalFace~\cite{Zhangzhang2011animal} dataset containing 20 classes of different animal faces. Although faces of different animals have their own unique characteristics, they still share a level of common structure. Therefore, instead of finding each class representation independently, it is reasonable to exploit the shared knowledge between all target classes. 
A basic but indirect example of knowledge sharing between the target classes can be seen in methods based on fine-tuning the pre-trained convolutional filters, where all target classes contribute to optimizing the same filters. The same, however, does not happen when learning the class-specific BN parameters independent of other classes.

\begin{figure}
\begin{center}
   \includegraphics[width=1\linewidth]{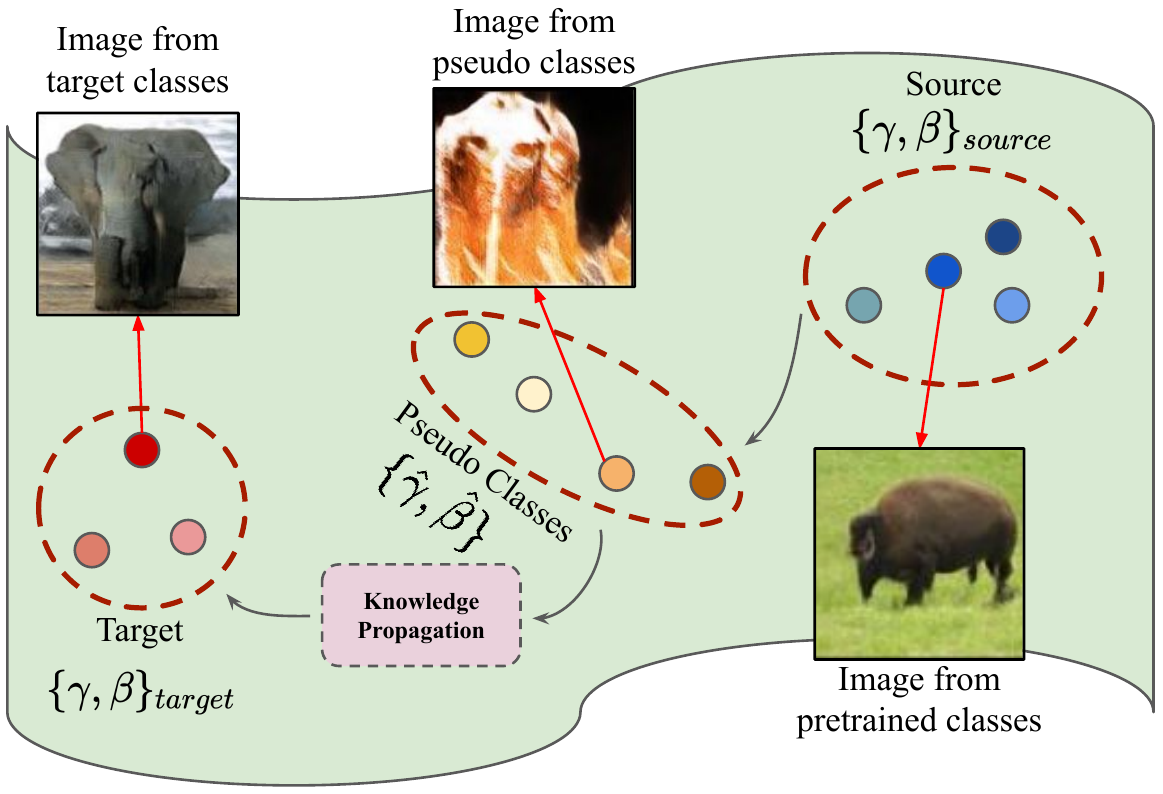}
\end{center}
   \caption{The visualization the proposed knowledge sharing and propagation. The sharing takes place in obtaining the pseudo-classes, by updating the source.
   The propagation step combines pseudo-classes to obtain the target. The sharing and the propagation are performed jointly. The images are obtained from our experiments (Please refer to Fig~\ref{fig:animal}-\ref{fig:ghosts}).}
\label{fig:manifold}
\end{figure}

To empower the knowledge sharing among target classes in our method, we propose to allow the target classes to jointly optimize the prior knowledge (pre-training BN parameters) during knowledge propagation. Optimizing the prior knowledge will enable the model to obtain a shared  set of intermediate representations, which are more suitable for all target classes. These shared representations--which we name pseudo-classes--then can be combined according to the similarity scores during the knowledge propagation step. Mathematically, we rewrite the propagation equation after knowledge sharing as:
\begin{equation}\label{eq:sharing}
\begin{split}
    &\gamma^{y'}_l = [\hat\gamma^1_l, \hat\gamma^2_l, \dots, \hat\gamma^N_l] {S^{y'}_{\gamma_l}}, \\
    &\beta^{y'}_l = [\hat\beta^1_l, \hat\beta^2_l, \dots, \hat\beta^N_l] {S^{y'}_{\beta_l}}, \\
    &y' \in \{N+1, \dots, N+M\}, \;\; l \in \{1, \dots, l\}.
\end{split}
\end{equation}
where $[\hat\gamma^y_l]$ and $[\hat\beta^y_l]$ are obtained by updating a copy of the source BN parameters, forming the shared representations for layer $l$. It is important to note that the modified BN parameters in ~\eqref{eq:sharing} are used only for learning the new classes.  We do not replace original old BN parameters, as we want to preserve the old knowledge during the transfer. This process does not further increase the number of parameters for inference, since the updated parameters (and the similarity scores) are discarded after learning the target BN parameters.  Fig.~\ref{fig:manifold} visualizes the proposed approach for knowledge sharing in conjunction with the knowledge propagation.
\begin{table*}[t]
\begin{center}
\scalebox{0.9}{
\begin{tabular}{|l|c|c|c|c|c|c||c|c|}
\hline
Method/ Experiment & 20/ 600 & 20/ 300 & 20/ 100 & 10/ 600 & 10/ 300 & 10/ 100 & mFID \\
\hline\hline
Scratch & \textbf{25.35}&	47.20&	81.90&	52.23&	62.17&	83.68&	58.76\\
\hline
TransferGAN & 27.67	&31.19&	48.67&	30.58&	40.74&	64.75&	40.60 \\
BSA - early & 37.12&	32.10&	50.88&	42.01&	40.34&	66.76&	44.87 \\
BSA - full & 28.03&	\textbf{30.89}&	40.73&	30.73&	35.62&	56.48&	37.08 \\
\hline
Ours & 29.10 & 30.99 &	\textbf{40.04} &	\textbf{29.95}&	\textbf{35.23}&	\textbf{54.95}&	\textbf{36.71} \\
\hline
\end{tabular}
}
\scalebox{0.9}{
\begin{tabular}{|l|c|c|}
\hline
Experiment & FID & KMMD\\
\hline\hline
Scratch* & 190 &  0.96\\
\hline
TransferGAN* & 89.2 & 0.53\\
PPGN* & 139 & 0.56\\
MineGAN* &  82.3 & 0.47\\
\hline
BSA** & 85.9 & 0.18\\
Ours** & \textbf{71.1} & \textbf{0.16}\\
\hline
\end{tabular}
}
\end{center}
\vspace{-0.4cm}
\caption{Results of the evaluation and the comparison of the proposed method using FID and KMMD score. Left: transferring 80 classes of CIFAR100 to 20 classes or 10 classes. Different columns report the best FID scores scores in difference cases. The format A/B means using A classes and B images per class for the target. BSA-early indicates the FID scores for BSA at the iteration where our method achieves its best FID score. Right: Results of transferring ImageNet to 5 classes of Places365 with 500 samples per class. \text{*} adopted from~\cite{wang2020minegan}. Some discrepancy between the KMMD scores of \text{*} and \text{*}\text{*} is possible\protect\footnotemark.}
\label{table:fid}
\end{table*}
\subsection{Training with Residuals and Sparsity}
For the purpose of generalizability, we also consider the classes that cannot be represented well only by combining the shared intermediate representations. To address this issue, we propose to add residual vectors $r^{y'}_{\gamma_l}$ and $r^{y'}_{\beta_l}$ to the scale and shift parameters obtained from knowledge propagation, respectively. However, in order to encourage the model to use the prior knowledge as much as possible, we minimize the magnitude of the residual vectors using $\ell_2$ regularization. Moreover, to encourage the new classes to learn from the most relevant prior knowledge, we also add an $\ell_1$ sparsity regularization on the similarity scores. 

\begin{figure}[t]
\begin{center}
   \includegraphics[width=0.85\linewidth]{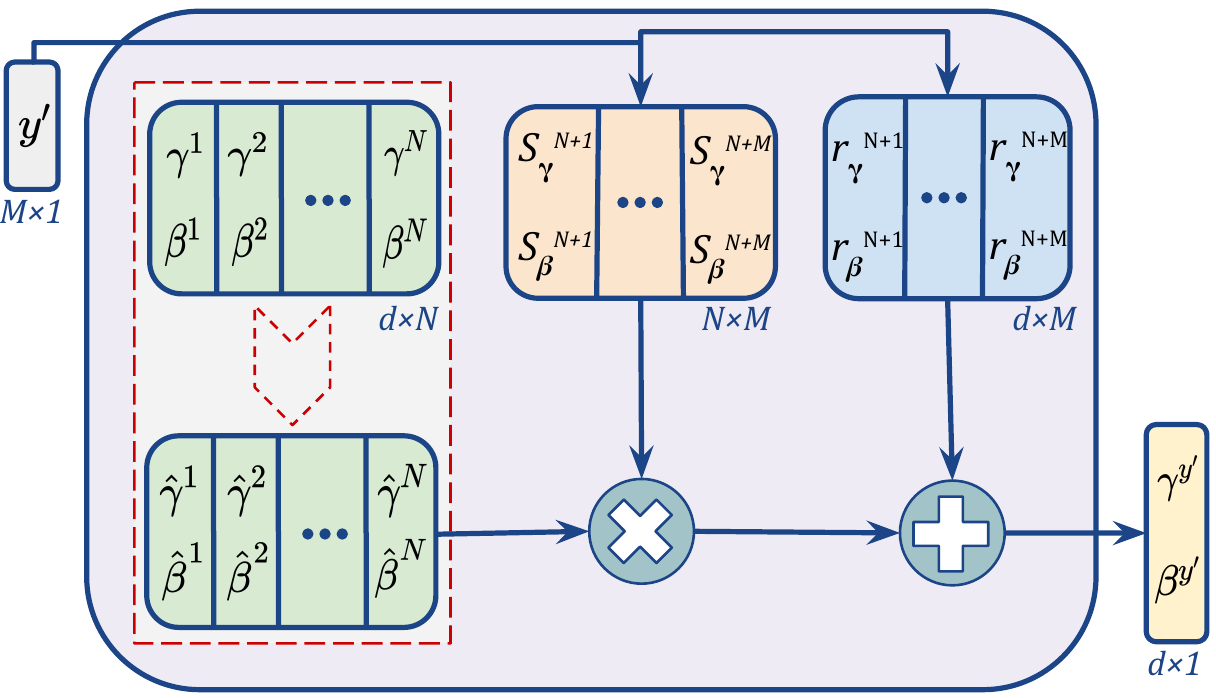}
\end{center}
   \caption{An overview of the proposed block for cGAN transfer with knowledge propagation across classes.}
\label{fig:block}
\end{figure}

In summary, we propose conditional GAN transfer from a pre-trained GAN to multi-class target data via BN parameters of pre-training classes, which are linearly combined to obtain the new classes' representations. Moreover, we enable sharing knowledge within the target classes by allowing them to update the prior knowledge according to the needs of the target data. An overview of the complete proposed method is illustrated in Fig.~\ref{fig:block}.

The final loss function of our method follows ~\eqref{eq:loss}, where $L_{g}$ is the adversarial loss~\cite{goodfellow2014gans, arjovsky2017wgan, lim2017geometric} used for training the GAN, $L_{r}$ is an $\ell_2$ regularization over the residuals, and $L_{s}$ is an $\ell_1$ regularization over the similarity scores:
\begin{equation}\label{eq:loss}
\begin{split}
    L =&\, L_{g} + \lambda_r L_{r} + \lambda_s L_{s}, \\
    =&\, L_{g} + \lambda_r \sum^L_{l=1}\sum^{N+M}_{y'=N+1}\{||r^{y'}_{\gamma_l}||^2_2 + ||r^{y'}_{\beta_l}||^2_2 \} \\
    &+ \lambda_s \sum^L_{l=1}\sum^{N+M}_{y'=N+1}\{|S^{y'}_{\gamma_l}| + |S^{y'}_{\beta_l}|\},
\end{split}
\end{equation}
where, $\lambda_r$ and $\lambda_s$ are the hyper-parameters associated to residual and sparsity losses, respectively.

\section{Experiments}
\footnotetext{Our choice of the width of the Gaussian kernel ($\sigma$) for calculating KMMD may differ from that of ~\cite{wang2020minegan}.}

In this section, we provide the details of our experiments, the evaluation of our method (cGANTransfer), and its comparison with the baseline methods in two different setups. Then, we provide further analysis of our contributions with an ablation study and explanatory visualizations.

\subsection{Experimental Setup}\label{sec5.1}
\noindent\textbf{Datasets:} To evaluate our method, we use two main experimental setups. In the first setup, we pre-train the network on 80 randomly-selected classes of CIFAR100~\cite{krizhevsky2009cifar}. The remaining classes are used as the target. To have a more thorough evaluation, we evaluate our method on CIFAR100 with different numbers of target classes and images per class. For the second setup, we consider the more challenging task of extending a network pre-trained on ImageNet~\cite{deng2009imagenet} to the Places365 dataset~\cite{zhou2014places}. Following MineGAN~\cite{wang2020minegan}, we select 5 classes (i.e., Alley, Arch, Art Gallery, Auditorium, Ballroom) and down-sample each class to 500 images. In addition, we use AnimalFace dataset~\cite{Zhangzhang2011animal}--containing 20 classes--for further analysis and visualizations. We down-sample each class to contain a maximum of 100 images.

\noindent\textbf{Architecture:} For the cGAN architecture, we use BigGAN~\cite{brock2018large} with hinge loss, as it is one of the most widely-used state-of-the-art cGANs. We use hierarchical noise for training on ImageNet, but not for CIFAR100, following the experimental setup of BigGAN~\cite{brock2018large}.

\noindent\textbf{Training:} To maintain our experiments in the class extension setup, we freeze the weights of the generator and only learn the parameters of our knowledge transfer block. We also find that, if the hierarchical noise is used (e.g. ImageNet setup), it is also necessary to fine-tune the linear layers that project the noise into the BN parameter's space. The reason might be that these linear layers are optimized to add the detailed style of the dataset to the generated images. 
Experiments on CIFAR100 are conducted using a single V100 GPU with a batch size of 50, and the experiments on the second setup are performed using 8 V100 GPUs with a batch size of 256.

\subsection{Comparison with the State-of-the-Art}\label{comparison}
To quantitatively evaluate our method, we use Fr\'echet Inception Distance (FID)~\cite{heusel2017fid}. We also provide the KMMD metric (Gaussian kernel with $\sigma=1$) for ImageNet setup, following BSA and MineGAN~\cite{noguchi2019image, wang2020minegan}. We compare our method against training from the scratch (Scratch), TransferGAN~\cite{wang2018transferring}, PPGN~\cite{nguyen2017PPGN}, MineGAN~\cite{wang2020minegan} and BSA~\cite{noguchi2019image}. Among these methods, BSA is the only one that performs the class extension. Therefore, we consider BSA as our main baseline.

\begin{figure}
\centering
   \includegraphics[height=4.5cm]{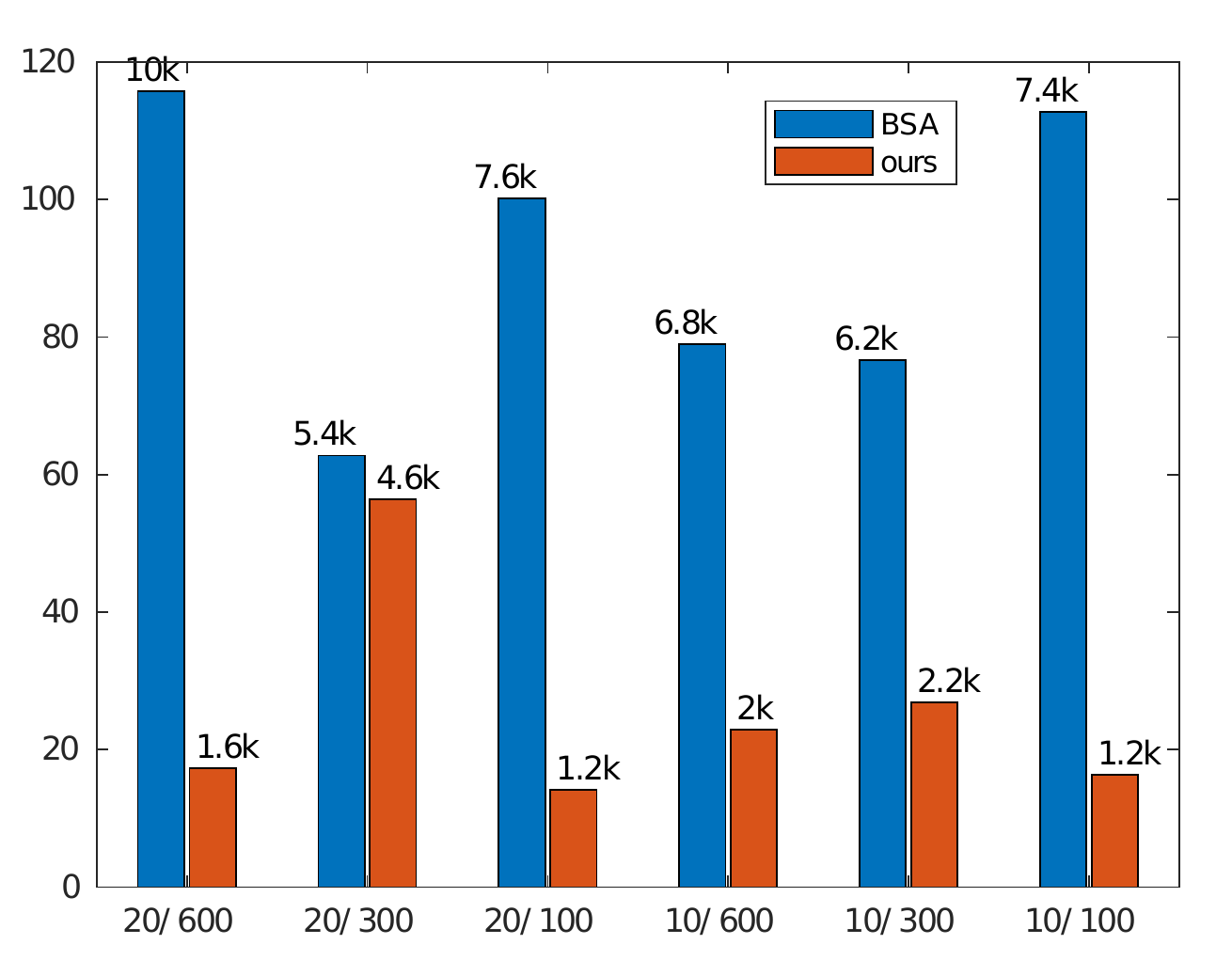}
   \includegraphics[height=4.3cm]{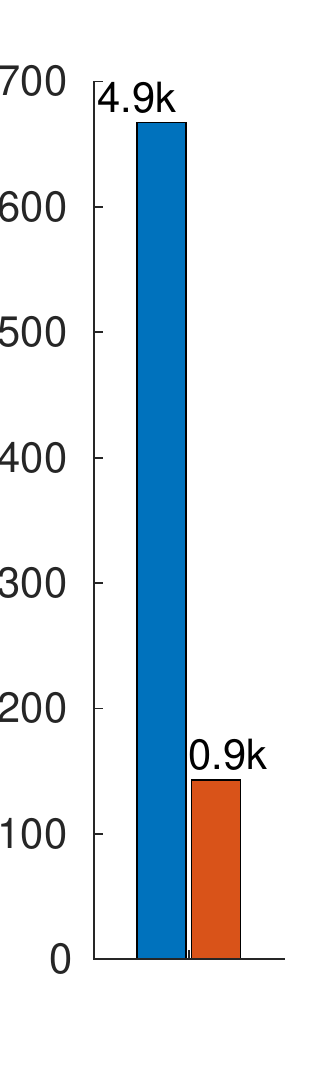}
\caption{The convergence comparison between BSA and ours on CIFAR (left) and Places365 (right) setups. Note that the number of iterations is shown above each bar.}
\label{fig:convergence}
\end{figure}

Table\ref{table:fid} (left) shows the results for evaluating our method on CIFAR100 and its comparison with the other methods. The experiments include two different numbers of classes (20 and 10), and 3 different numbers of samples per class (600, 300, 100). As it can be seen, decreasing the number of training samples degrades the performance of learning from scratch and TransferGAN significantly. On the other hand, BSA and our method perform more robustly on small data. Compared to BSA, our method achieves slightly better FID scores (possibly because of the less challenging task on CIFAR) and a significantly better convergence speed (more than \textbf{4x} speed-up on average), as visualized in Fig.~\ref{fig:convergence} (left). Our method enjoys such speed-up thanks to the knowledge propagation through the class representations, as opposed to its counterpart--BSA.
\begin{figure}
\begin{center}
   \includegraphics[width=\linewidth]{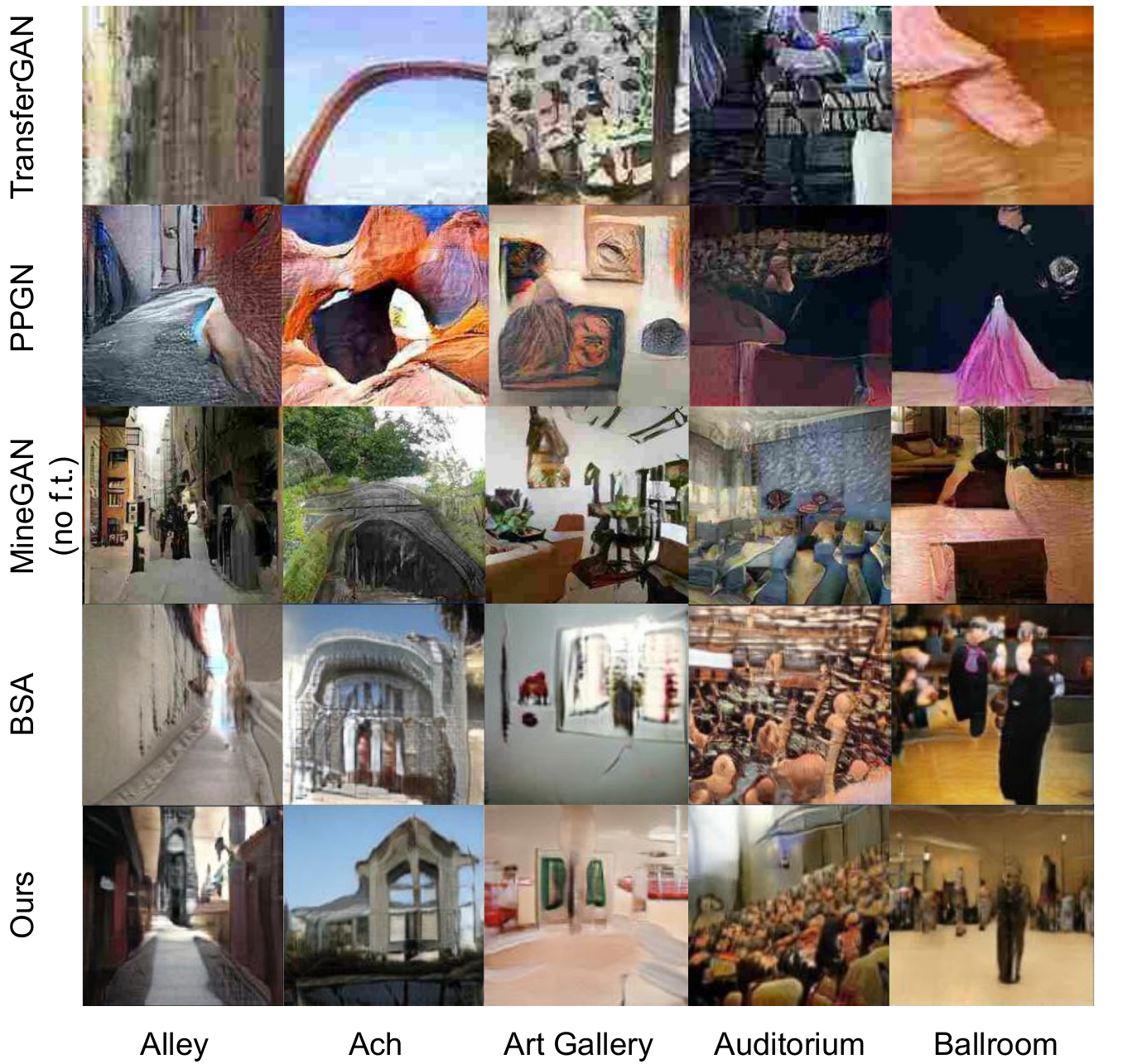}
\end{center}
 \vspace{-0.4cm}
   \caption{The visual comparison between our proposed method and baseline methods on Places365. The first three Competitors’ results are adopted from~\cite{wang2020minegan}.}
\label{fig:places}
\end{figure}
The superiority of the proposed method becomes even more noticeable in the more challenging setup of transferring ImageNet to Places365. The classes of Places365 have large intra-class diversity, and there is a considerable domain gap between the source and target datasets. Therefore, the task becomes highly challenging when only use 500 images per class are used. Table~\ref{table:fid} (right) presents the results for transferring ImageNet to 5 classes of Places365. As it can be seen, our method clearly outperforms the other methods in terms of FID and KMMD. Fig.~\ref{fig:convergence} (right) again shows that the proposed method is more than \textbf{5x} faster than BSA in terms of convergence speed. The comparison between the results of our method and BSA shows that knowledge transfer across classes results in not only faster convergence but also a more accurate distribution approximation of the target data. Fig.~\ref{fig:places} depicts the visual results for our method and the compared ones in ImageNet-to-Places365 setup.
\subsection{Ablation Study}
To show the effect of each component in our method, we perform the following experiments on ImageNet-to-Places365 setup: Transferring from previous classes without updating the prior knowledge or residual learning, transferring from updatable prior knowledge without residual learning, updatable prior with residuals, sharing the combination weights between the layers, and the experiments on the regularizations. Table~\ref{table:ablation} shows the results for the ablation study of our method when transferring from ImageNet to Places. The ablation study shows that, all the proposed components contribute meaningfully to the final model.

\begin{table}
\begin{center}
\scalebox{0.9}{
\begin{tabular}{|l|c|c|}
\hline
Experiment & Best FID & Iterations\\
\hline\hline
Freezed Prior, w/o Res & 96.3 & 3000 \\
Freezed Prior, with Res & 79.4 & 1100 \\
Tunable Prior, w/o Res & 81.2 & 4000 \\
Shared W, Tunable Prior, w/o Res & 79.8 & 3400 \\
Final architecture w/o reg & 82.4 & 2400\\
Final architecture w/o l1  & 84.7 & 1600 \\
Final architecture w/o l2 & 81.1 & 2100 \\
Final architecture, with l1 \& l2 & \textbf{71.1}  & \textbf{900} \\
\hline
\end{tabular}
}
\end{center}
\vspace{-0.2cm}
\caption{Ablation study over GAN transfer from ImageNet to Places365. Iterations: iteration number for the best FID.}
\label{table:ablation}
\end{table}
\subsection{Fine-tuning}
As mentioned in Sec.~\ref{sec5.1}, we freeze the weights of the generator in our experiments. This enables a parameter-efficient extension of the cGAN to target classes. However, when parameter efficiency is not a constraint, fine-tuning the filters could further improve the performance of the model, as shown in~\cite{wang2020minegan}. Therefore, we also provide the results of further fine-tuning of our model after learning the target BN parameters in Table~\ref{table:ft}. To fine-tune the model, we freeze the knowledge propagation parameters (prior knowledge and similarity scores) and only fine-tune the residuals, as well as the rest of the model. The results show the additional benefit of further fine-tuning of our model.
\begin{table}[h]
\begin{center}
\scalebox{0.9}{
\begin{tabular}{|l|c|c|}
\hline
Dataset & FID & KMMD\\
\hline\hline
Places365  & 65.48 &  0.156 \\
AnimalFace & 68.92 & 0.194 \\

\hline
\end{tabular}
}
\end{center}
\vspace{-0.2cm}
\caption{The results of further fine-tuning of our model on Places365 and AnimalFace.}
\label{table:ft}
\end{table}
\subsection{Analysis and Visualizations}
\begin{figure}
\begin{center}
\begin{tabular}{|l|c|c|c|c|}
\hline
Experiment & FID & KMMD & Iters & Time (Mins) \\
\hline\hline
BSA & 91.9 & 0.25 & 4800 & 650 \\
Ours & \textbf{85.9} & \textbf{0.23} & \textbf{1400} & \textbf{235}\\
\hline
\end{tabular}
\end{center} 
\vspace{-0.5cm}
\begin{center}
   \includegraphics[width=\linewidth]{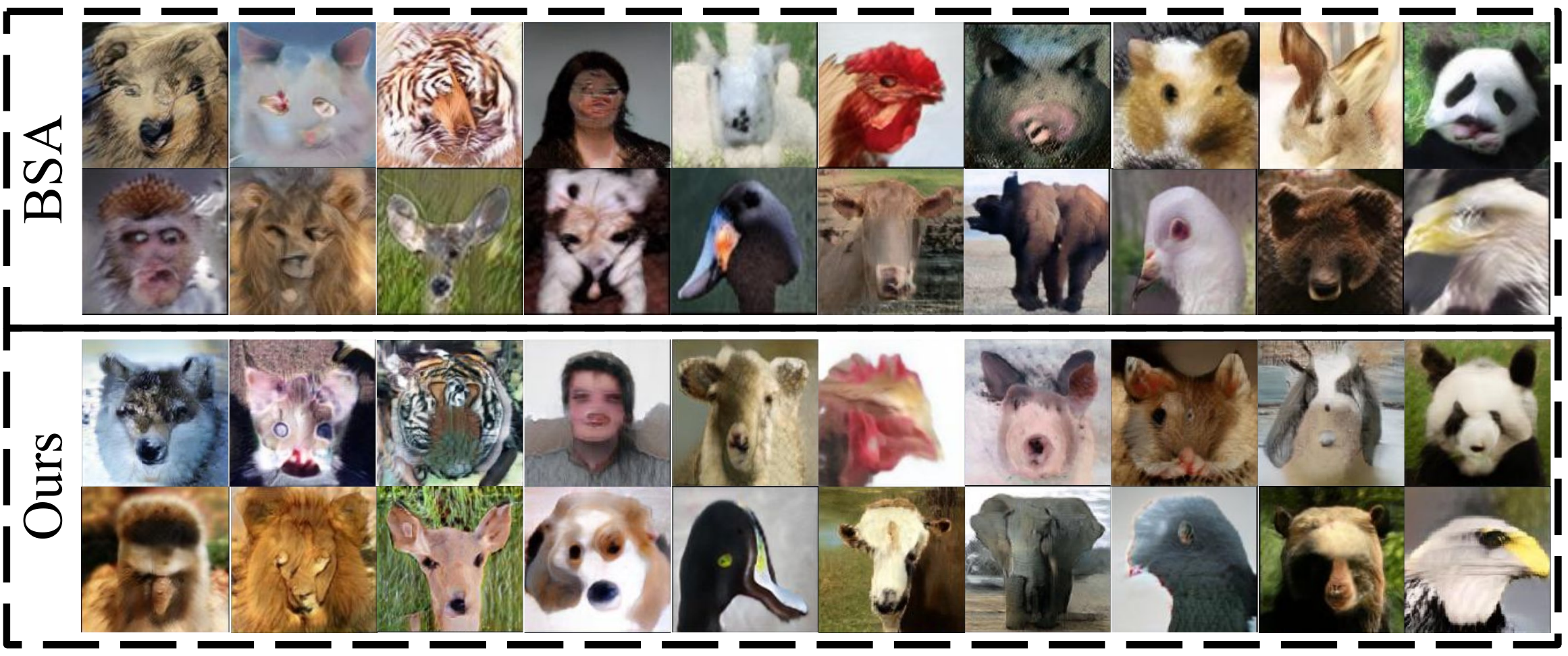}
\end{center}
\vspace{-0.3cm}
\caption{Quantitative (top) and qualitative (bottom) results of transferring ImageNet to 20 classes of Face Animal without (BSA) and with (Ours) \textbf{cross-class knowledge propagation}. Iters: training iteration number for the best FID.}
\label{fig:animal}
\end{figure}

To better understand the proposed method, we further analyze its main components while transferring ImageNet to AnimalFace. The quantitative (FID, KMMD, and convergence time) and qualitative results of class extension to AnimalFace, with and without class knowledge propagation, are shown in Fig.~\ref{fig:animal}. The results for AnimalFace are consistent with the previous results provided in Sec.~\ref{comparison}.

\begin{figure}
\vspace{-0.3cm}
\begin{center}
   \includegraphics[width=\linewidth]{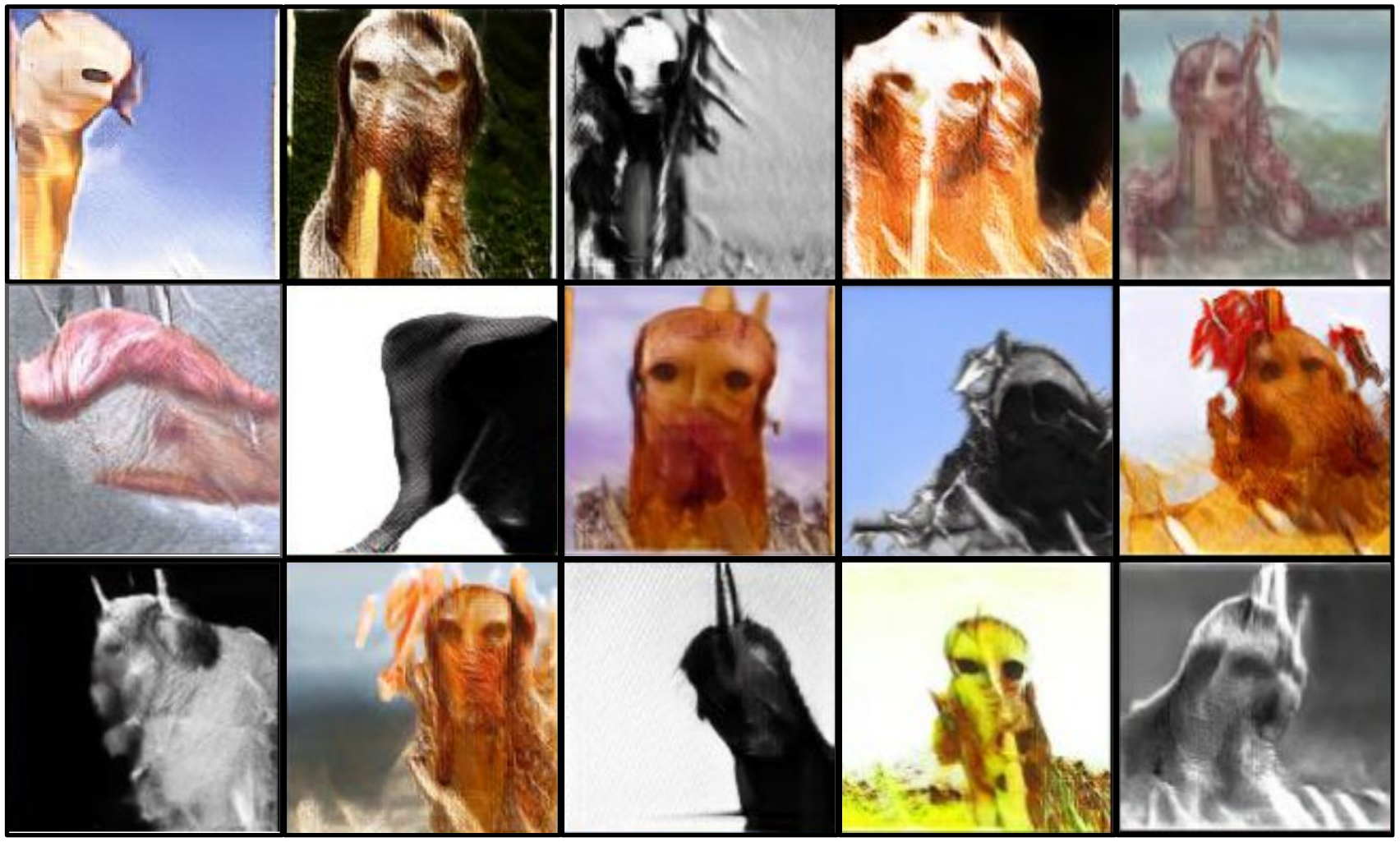}
\end{center}
\vspace{-0.3cm}
   \caption{\textbf{Knowledge sharing.} The intermediate shared representations learned for AnimalFace as the target data.}
\label{fig:ghosts}
\end{figure}

\noindent\textbf{Knowledge sharing within target classes:} An important component of our method is knowledge sharing within the target classes.  Fig.~\ref{fig:ghosts}. shows the pseudo-classes (intermediate representations) obtained by our method after training on the target data. To visualize these representations, we directly sample the BN parameters from the updated base representations, instead of combining them to obtain the new class representations (refer to Fig.~\ref{fig:manifold}). As it can be seen, the initial class representations of the ImageNet are transformed to some intermediate ``pseudo-class" representations that contain the shared structure of the target faces, but do not belong to a particular target class. Thus, our method is able to obtain the representations of each new class by combining these new pseudo-classes, which are learned using all of the target classes.

\begin{figure}
\vspace{-0.3cm}
\begin{center}
   \includegraphics[width=\linewidth]{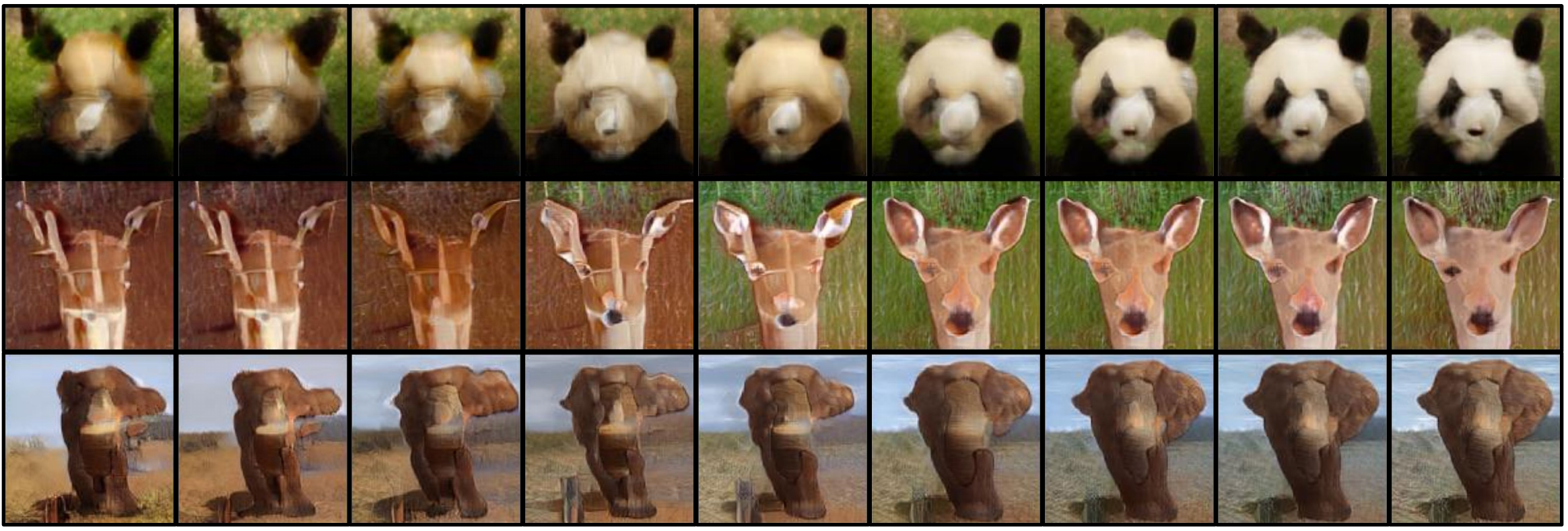}
\end{center}
\vspace{-0.3cm}
   \caption{\textbf{Layer-wise knowledge transfer.} Different layers introduce different information to the generated images. The shape is mostly changed by the starting layers, while colors and finer details are added by the later layers.}
\label{fig:spectrum}
\end{figure}

\noindent\textbf{Layer-specific knowledge transfer:} Another key aspect of our method, as discussed in Sec.~\ref{sec:method} (refer to Fig.~\ref{fig:layerwise}) and supported by the ablation study (Table~\ref{table:ablation}), is that the linear transfer weights are different for each layer. Fig.~\ref{fig:spectrum} visualizes the effect of using different weights per layer. To obtain these visualizations, after training the network on the target, we use the learned similarity scores of the first layer for all the layers. Then, we gradually introduce the learned scores of the next layers. From the results, we see that the first layers are mostly responsible for the general object shape, and the later layers introduce color and finer details.

\section{Conclusion and Future Work}
In this paper, we studied the problem of conditional GAN transfer by transferring the knowledge across both source and target classes. We represented the knowledge of individual classes by their respective batch normalization parameters, which are used for conditioning during the generation. 
To propagate the knowledge to new classes, we introduced a method that learns to update and combine the batch normalization parameters of the source classes. 
The evaluations on three standard benchmarks demonstrate a clear advantage of our method, both in terms of training efficiency and the image generation quality (in terms of FID and KMMD), compared to the state-of-the-art methods. 
Our ablation study showed the importance of jointly using the update and combination steps, which we referred to as sharing and propagation, respectively.

We believe our study can be followed by several interesting future works, such as further knowledge propagation on the discriminator, its integration with differentiable augmentation~\cite{Karras2020ada, zhao2020differentiable}, and the extension to lifelong learning.
\section*{Acknowledgements}
This work was partly supported by EU Horizion 2020 Programme ENCORE (grant no.\ 820434). Valuable computing resources were provided by Amazon AWS.

{\small
\bibliographystyle{ieee_fullname}
\bibliography{cvpr}

\begin{thebibliography}{10}\itemsep=-1pt

\bibitem{arjovsky2017wgan}
Martin Arjovsky, Chintala, Soumith, and Léon Bottou.
\newblock Wasserstein generative adversarial networks.
\newblock {\em arXiv preprint arXiv:1701.07875}, 2017.

\bibitem{bengio2009curriculum}
Yoshua Bengio, J{\'e}r{\^o}me Louradour, Ronan Collobert, and Jason Weston.
\newblock Curriculum learning.
\newblock In {\em Proceedings of the 26th annual international conference on
  machine learning}, pages 41--48, 2009.

\bibitem{brock2018large}
Andrew Brock, Jeff Donahue, and Karen Simonyan.
\newblock Large scale gan training for high fidelity natural image synthesis.
\newblock {\em arXiv preprint arXiv:1809.11096}, 2018.

\bibitem{clark2019adversarial}
Aidan Clark, Jeff Donahue, and Karen Simonyan.
\newblock Adversarial video generation on complex datasets.
\newblock {\em arXiv preprint arXiv:1907.06571}, 2019.

\bibitem{cong2020ganmemory}
Yulai Cong, Miaoyun Zhao, Jianqiao Li, Sijia Wang, and Lawrence Carin.
\newblock Gan memory with no forgetting.
\newblock In {\em Advances in Neural Information Processing Systems}, pages
  469--477, 2020.

\bibitem{data2018interpolating}
Gratianus W.~P. Data, Kirjon Ngu, David~W. Murray, and Victor~A. Prisacariu.
\newblock Interpolating convolutional neural networks using batch
  normalization.
\newblock In {\em Proceedings of the European Conference on Computer Vision},
  2018.

\bibitem{devries2017modulating}
Harm de Vries, Florian Strub, Jérémie Mary, Hugo Larochelle, Olivier
  Pietquin, and Aaron Courville.
\newblock Modulating early visual processing by language.
\newblock In {\em Advances in Neural Information Processing Systems}, pages
  6594--6604, 2017.

\bibitem{deng2009imagenet}
Jia Deng, Wei Dong, Richard Socher, Li-Jia Li, Kai Li, and Li Fei-Fei.
\newblock Imagenet: A large-scale hierarchical image database.
\newblock In {\em IEEE conference on computer vision and pattern recognition},
  pages 248--255, 2009.

\bibitem{ganin2015unsupervised}
Yaroslav Ganin and Victor Lempitsky.
\newblock Unsupervised domain adaptation by backpropagation.
\newblock In {\em International conference on machine learning}, pages
  1180--1189. PMLR, 2015.

\bibitem{goodfellow2014gans}
Ian Goodfellow, Jean Pouget-Abadie, Mehdi Mirza, Bing Xu, David Warde-Farley,
  Sherjil Ozair, Aaron Courville, and Yoshua Bengio.
\newblock Generative adversarial nets.
\newblock In {\em Advances in Neural Information Processing Systems}, pages
  2672--2680, 2014.

\bibitem{guo2017active}
Yuchen Guo, Guiguang Ding, Yue Gao, and Jungong Han.
\newblock Active learning with cross-class similarity transfer.
\newblock In {\em Association for the Advancement of Artificial Intelligence},
  pages 1338--1344, 2017.

\bibitem{heusel2017fid}
Martin Heusel, Hubert Ramsauer, Thomas Unterthiner, Bernhard Nessler, and Sepp
  Hochreiter.
\newblock Gans trained by a two time-scale update rule converge to a local nash
  equilibrium.
\newblock In {\em Advances in Neural Information Processing Systems}, pages
  6626--6637, 2017.

\bibitem{hinton2015distilling}
Geoffrey Hinton, Oriol Vinyals, and Jeff Dean.
\newblock Distilling the knowledge in a neural network.
\newblock {\em arXiv preprint arXiv:1503.02531}, 2015.

\bibitem{karras2017progressive}
Tero Karras, Timo Aila, Samuli Laine, and Jaakko Lehtinen.
\newblock Progressive growing of gans for improved quality, stability, and
  variation.
\newblock In {\em International Conference on Learning Representations}, 2018.

\bibitem{Karras2020ada}
Tero Karras, Miika Aittala, Janne Hellsten, Samuli Laine, Jaakko Lehtinen, and
  Timo Aila.
\newblock Training generative adversarial networks with limited data.
\newblock In {\em Advances in Neural Information Processing Systems}, 2020.

\bibitem{karras2019style}
Tero Karras, Samuli Laine, and Timo Aila.
\newblock A style-based generator architecture for generative adversarial
  networks.
\newblock In {\em Proceedings of the IEEE conference on computer vision and
  pattern recognition}, pages 4401--4410, 2019.

\bibitem{kavalerov2019cgan}
Ilya Kavalerov, Wojciech Czaja, and Rama Chellappa.
\newblock cgans with multi-hinge loss.
\newblock {\em arXiv preprint arXiv:1912.04216}, 2019.

\bibitem{krizhevsky2009cifar}
Alex Krizhevsky.
\newblock Learning multiple layers of features from tiny images.
\newblock {\em Tech Report}, 2009.

\bibitem{li2017learning}
Zhizhong Li and Derek Hoiem.
\newblock Learning without forgetting.
\newblock {\em IEEE transactions on pattern analysis and machine intelligence},
  40(12):2935--2947, 2017.

\bibitem{lim2017geometric}
Jae~Hyun lim and Jong~Chul Ye.
\newblock Geometric gan.
\newblock {\em arXiv preprint arXiv:1705.02894}, 2017.

\bibitem{liu2020generative}
Ming-Yu Liu, Xun Huang, Jiahui Yu, Ting-Chun Wang, and Arun Mallya.
\newblock Generative adversarial networks for image and video synthesis:
  Algorithms and applications.
\newblock {\em arXiv preprint arXiv:2008.02793}, 2020.

\bibitem{liu2016coupled}
Ming-Yu Liu and Oncel Tuzel.
\newblock Coupled generative adversarial networks.
\newblock In {\em Advances in Neural Information Processing Systems}, pages
  469--477, 2016.

\bibitem{liu2020open}
Ziwei Liu, Zhongqi Miao, Xingang Pan, Xiaohang Zhan, Dahua Lin, Stella~X Yu,
  and Boqing Gong.
\newblock Open compound domain adaptation.
\newblock In {\em Proceedings of the IEEE/CVF Conference on Computer Vision and
  Pattern Recognition}, 2020.

\bibitem{mirza2014cgan}
Mehdi Mirza and Simon Osindero.
\newblock Conditional generative adversarial nets.
\newblock {\em arXiv preprint arXiv:1411.1784}, 2014.

\bibitem{miyato2018spectralnorm}
Takeru Miyato, Toshiki Kataoka, Masanori Koyama, and Yuichi Yoshida.
\newblock cgans with projection discriminator.
\newblock {\em arXiv preprint arXiv:1802.05957}, 2018.

\bibitem{miyato2018cganproj}
Takeru Miyato and Masanori Koyama.
\newblock cgans with projection discriminator.
\newblock {\em arXiv preprint arXiv:1802.05637}, 2018.

\bibitem{mo2020freeze}
Sangwoo Mo, Minsu Cho, and Jinwoo Shin.
\newblock Freeze discriminator: A simple baseline for fine-tuning gans.
\newblock {\em arXiv preprint arXiv:2002.10964}, 2020.

\bibitem{nguyen2017PPGN}
Anh Nguyen, Jeff Clune, Yoshua Bengio, Alexey Dosovitskiy, and Jason Yosinski.
\newblock Plug \& play generative networks: Conditional iterative generation of
  images in latent space.
\newblock In {\em Proceedings of the IEEE Conference on Computer Vision and
  Pattern Recognition}, pages 4467--4477, 2017.

\bibitem{nguyen2016DGNAM}
Anh Nguyen, Alexey Dosovitskiy, Jason Yosinski, Thomas Brox, and Jeff Clune.
\newblock Synthesizing the preferred inputs for neurons in neural networks via
  deep generator networks.
\newblock In {\em Advances in Neural Information Processing Systems}, pages
  3387--3395, 2016.

\bibitem{noguchi2019image}
Atsuhiro Noguchi and Tatsuya Harada.
\newblock Image generation from small datasets via batch statistics adaptation.
\newblock In {\em Proceedings of the IEEE International Conference on Computer
  Vision}, pages 2750--2758, 2019.

\bibitem{pan2009survey}
Sinno~Jialin Pan and Qiang Yang.
\newblock A survey on transfer learning.
\newblock {\em IEEE Transactions on knowledge and data engineering},
  22(10):1345--1359, 2009.

\bibitem{panareda2017open}
Pau Panareda~Busto and Juergen Gall.
\newblock Open set domain adaptation.
\newblock In {\em Proceedings of the IEEE International Conference on Computer
  Vision}, pages 754--763, 2017.

\bibitem{parisi2019continual}
German~I Parisi, Ronald Kemker, Jose~L Part, Christopher Kanan, and Stefan
  Wermter.
\newblock Continual lifelong learning with neural networks: A review.
\newblock {\em Neural Networks}, 113:54--71, 2019.

\bibitem{park2019semantic}
Taesung Park, Ming-Yu Liu, Ting-Chun Wang, and Jun-Yan Zhu.
\newblock Semantic image synthesis with spatially-adaptive normalization.
\newblock In {\em Proceedings of the IEEE Conference on Computer Vision and
  Pattern Recognition}, pages 2337--2346, 2019.

\bibitem{Zhangzhang2011animal}
Zhangzhang Si and Song-Chun Zhu.
\newblock Learning hybrid image templates (hit) by information projection.
\newblock {\em IEEE Transactions on pattern analysis and machine intelligence},
  34(7):1354--1367, 2011.

\bibitem{sung2018learning}
Flood Sung, Yongxin Yang, Li Zhang, Tao Xiang, Philip~HS Torr, and Timothy~M
  Hospedales.
\newblock Learning to compare: Relation network for few-shot learning.
\newblock In {\em Proceedings of the IEEE Conference on Computer Vision and
  Pattern Recognition}, pages 1199--1208, 2018.

\bibitem{tzeng2017adversarial}
Eric Tzeng, Judy Hoffman, Kate Saenko, and Trevor Darrell.
\newblock Adversarial discriminative domain adaptation.
\newblock In {\em Proceedings of the IEEE conference on computer vision and
  pattern recognition}, pages 7167--7176, 2017.

\bibitem{vinyals2016matching}
Oriol Vinyals, Charles Blundell, Timothy Lillicrap, Daan Wierstra, et~al.
\newblock Matching networks for one shot learning.
\newblock In {\em Advances in Neural Information Processing Systems}, pages
  3630--3638, 2016.

\bibitem{wang2020minegan}
Yaxing Wang, Abel Gonzalez-Garcia, David Berga, Luis Herranz, Fahad~Shahbaz
  Khan, and Joost van~de Weijer.
\newblock Minegan: effective knowledge transfer from gans to target domains
  with few images.
\newblock In {\em Proceedings of the IEEE/CVF Conference on Computer Vision and
  Pattern Recognition}, pages 9332--9341, 2020.

\bibitem{wang2018transferring}
Yaxing Wang, Chenshen Wu, Luis Herranz, Joost van~de Weijer, Abel
  Gonzalez-Garcia, and Bogdan Raducanu.
\newblock Transferring gans: generating images from limited data.
\newblock In {\em Proceedings of the European Conference on Computer Vision},
  pages 218--234, 2018.

\bibitem{wu2019sliced}
Jiqing Wu, Zhiwu Huang, Dinesh Acharya, Wen Li, Janine Thoma, Danda~Pani
  Paudel, and Luc~Van Gool.
\newblock Sliced wasserstein generative models.
\newblock In {\em Proceedings of the IEEE/CVF Conference on Computer Vision and
  Pattern Recognition}, pages 3713--3722, 2019.

\bibitem{zamir2018taskonomy}
Amir~R Zamir, Alexander Sax, William Shen, Leonidas~J Guibas, Jitendra Malik,
  and Silvio Savarese.
\newblock Taskonomy: Disentangling task transfer learning.
\newblock In {\em Proceedings of the IEEE conference on computer vision and
  pattern recognition}, pages 3712--3722, 2018.

\bibitem{zhang2019sagan}
Han Zhang, Ian Goodfellow, Dimitris Metaxas, and Augustus Odena.
\newblock Self-attention generative adversarial networks.
\newblock In {\em International conference on machine learning}, pages
  7354--7363. PMLR, 2019.

\bibitem{zhao2020leveraging}
Miaoyun Zhao, Yulai Cong, and Lawrence Carin.
\newblock On leveraging pretrained gans for limited-data generation.
\newblock {\em arXiv preprint arXiv:2002.11810}, 2020.

\bibitem{zhao2020differentiable}
Shengyu Zhao, Zhijian Liu, Ji Lin, Jun-Yan Zhu, and Song Han.
\newblock Differentiable augmentation for data-efficient gan training.
\newblock In {\em Advances in Neural Information Processing Systems},
  volume~33, 2020.

\bibitem{zhou2014places}
Bolei Zhou, Agata Lapedriza, Jianxiong Xiao, Antonio Torralba, and Aude Oliva.
\newblock Learning deep features for scene recognition using places database.
\newblock In {\em Advances in Neural Information Processing Systems}, pages
  487--495, 2014.

\bibitem{zhou2020domain}
Fan Zhou, Zhuqing Jiang, Changjian Shui, Boyu Wang, and Brahim Chaib-draa.
\newblock Domain generalization with optimal transport and metric learning.
\newblock {\em arXiv preprint arXiv:2007.10573}, 2020.

\end{thebibliography}
}

\renewcommand\thesection{\Alph{section}}

\newpage


\appendix
\section*{Supplementary Material}

First, we provide more details on the model architecture and implementation of our experimental setup. Then, we further discuss the quantitative results of CIFAR100~\cite{krizhevsky2009cifar} experiments. Moreover, we discuss source-to-target similarities, training curves, and single-class target data. Finally, we provide more visual results obtained using our proposed method.

\section{Additional Implementation Details}
\begin{figure}
\begin{center}
   \scalebox{0.5}{\includegraphics[width=\linewidth]{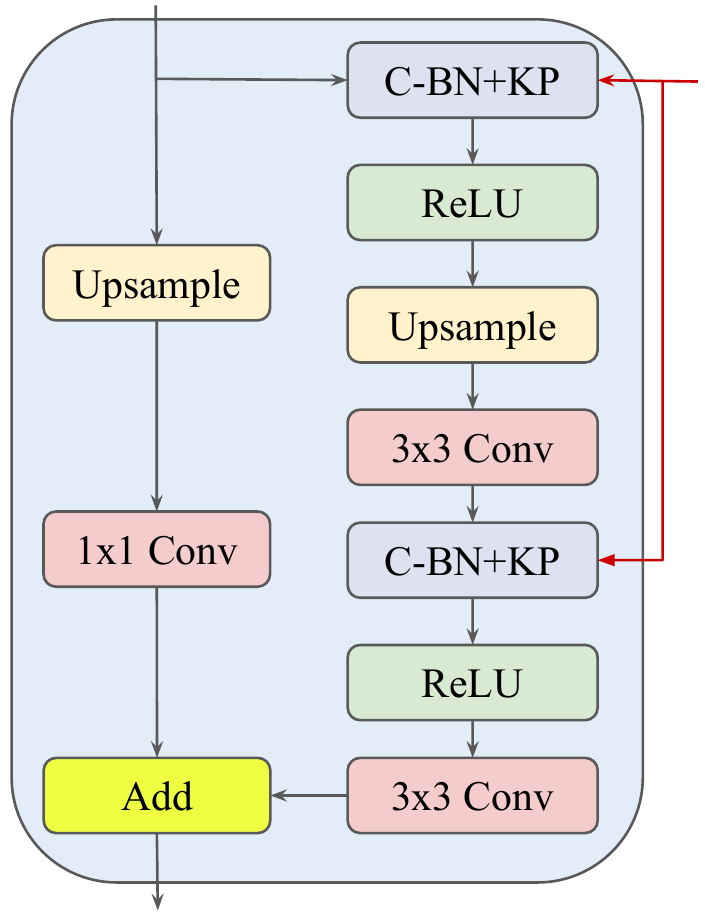}}
   \scalebox{0.46}{\includegraphics[width=\linewidth]{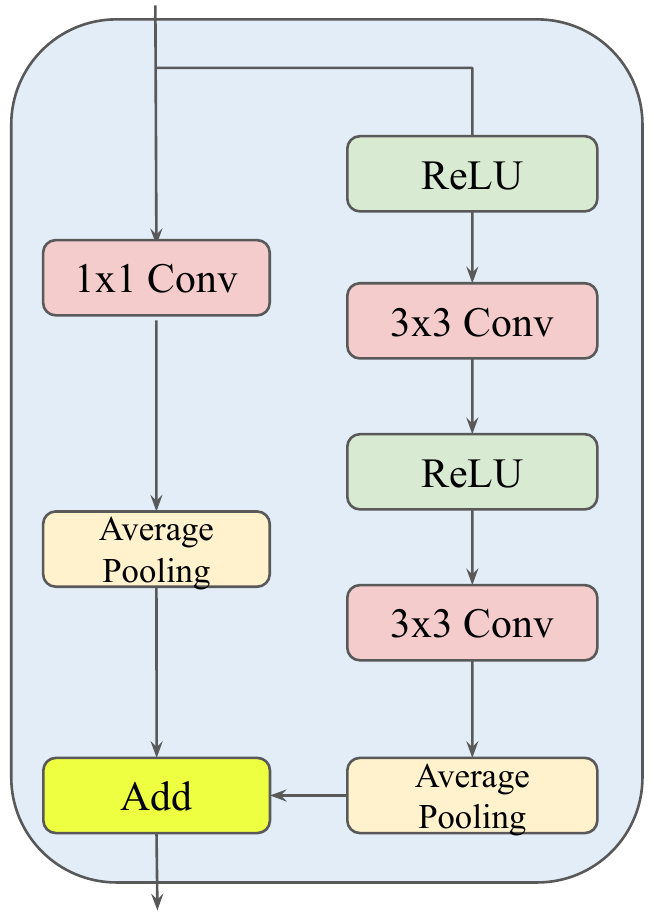}}
\end{center}
\vspace{-0.4cm}
   \caption{The architecture of the residual blocks used in the network. Left: the generator's ResBlock (``C-BN+KP" layer indicates the conditional batch normalization with knowledge propagation across classes. See Fig. 4 of the main paper for more details). Right: the discriminator's ResBlock.}
\label{fig:res_block}
\end{figure}

\noindent\textbf{Architecture:} In the main paper, we follow \cite{noguchi2019image, wang2020minegan} to employ the architecture of BigGAN~\cite{brock2018large} as our backbone for the GAN transfer tasks. It is worth mentioning that, the BigGAN implementation for CIFAR uses the basic form of BigGAN, which for example, does not use a hierarchical latent variable. In particular, Tables \ref{table:cifar_arch} and \ref{table:imagenet_arch} show the network architecture for CIFAR100 and ImageNet~\cite{deng2009imagenet} setups. Fig. \ref{fig:res_block} shows the architecture of the residual blocks used in the generator and the discriminator. The detailed diagram for the conditional batch normalization layer with knowledge propagation across classes has been provided in Fig. 4 of the main paper.

\noindent\textbf{Baselines:} The original study proposing batch normalization adaptation (BSA)~\cite{noguchi2019image} uses supervised loss function (L1/Perceptual loss) instead of adversarial loss. In our experiments, whenever we refer to BSA, we mean training the GAN model adversarially, while freezing the filters and learning the BN parameters from the scratch. Therefore our implementation of BSA could be considered as the main baseline for our experiments since it shares the same setup as ours, but without performing any knowledge transfer across the classes.

\noindent\textbf{Ablation Study Experiments:} Table 2 in the main paper shows the results for the ablation study on the ImageNet-to-Places365 setup. In the table, ``Prior" refers to the BN parameters of the previous classes. The table includes the results for using the prior with and without being further updated via knowledge sharing using target classes. ``Shared W" in the fourth experiment refers to using shared similarity wights over all layers of the generator to combine previous BN parameters of each layer. The term ``w/o reg" in the fifth experiment refers to not using l1 regularization on the combination weights and l2 regularization on the residuals.
\begin{table}
\begin{center}
\scalebox{0.85}{
\begin{tabular}{c}
\hline
\hline
$z \in {\mathbb{R}^{128}} \sim \mathcal{N}(0, I)$ \\
\hline
Linear $4 \times 4 \times 256$ \\
\hline
ResBlock up 256 \\
\hline
ResBlock up 256 \\
\hline
ResBlock up 256 \\
\hline
BN, ReLU, Conv $3\times3$, Tanh \\
\hline
\hline
\end{tabular}
}
\scalebox{0.85}{
\begin{tabular}{c}
\hline
\hline
$x \in \mathbb{R}^{32\times32\times3}$ \\
\hline
ResBlock down 64\\
\hline
ResBlock down 128\\
\hline
ResBlock down 256\\
\hline
ResBlock down 512\\
\hline
ResBlock 1024 \\
\hline
ReLU, Global Sum Pooling \\
\hline
Embed(y).h + (Linear→1) \\
\hline
\hline
\end{tabular}
}
\end{center}
\vspace{-0.4cm}
\caption{The network architecture for CIFAR setup: Left: the generator. Right: the discriminator.}
\label{table:cifar_arch}
\end{table}
\begin{table}
\begin{center}
\scalebox{0.85}{
\begin{tabular}{c}
\hline
\hline
$z \in {\mathbb{R}^{120}} \sim \mathcal{N}(0, I)$ \\
\hline
Linear $4 \times 4 \times 256$ \\
\hline
ResBlock up 256 \\
\hline
ResBlock up 256 \\
\hline
ResBlock up 256 \\
\hline
BN, ReLU, Conv $3\times3$, Tanh \\
\hline
\hline
\end{tabular}
}
\scalebox{0.85}{
\begin{tabular}{c}
\hline
\hline
$x \in \mathbb{R}^{128\times128\times3}$ \\
\hline
ResBlock down 96\\
\hline
None-Local Block ($64\times64$) \\
\hline
ResBlock down 192 \\
\hline
ResBlock down 384 \\
\hline
ResBlock down 768\\
\hline
ResBlock down 1536\\
\hline
ResBlock 1536 \\
\hline
ReLU, Global Sum Pooling \\
\hline
Embed(y).h + (Linear→1) \\
\hline
\hline
\end{tabular}
}
\end{center}
\vspace{-0.4cm}
\caption{The network architecture for ImageNet setup: Left: the generator. Right: the discriminator.}
\label{table:imagenet_arch}
\end{table}
\section{Further Discussion on Quantitative Results}
In Table 1 of the main paper, which shows the FID scores for different experiments on CIFAR100, the results of the first experiment (20 classes, 600 samples per class) is marginally different from those of the other experiments. It can be seen that, learning from the scratch performs better than all of the transfer learning methods, since the training data is large enough for learning the filters from scratch. However, by reducing the sample number in the next experiments, the performance of learning from scratch immediately deteriorates, while the transfer learning methods remain more robust. However, after reducing the training data even more (20/100, 10/600, 10/300, 10/100), fine-tuning (TransferGAN~\cite{wang2018transferring}) also degrades significantly compared to BSA and our method, and it falls into mode collapse. Comparing the FID scores of BSA and our method on CIFAR, although comparable, it can be observed that our method starts to perform better in the experiments with less amount of data, showing the importance of using prior knowledge from previous classes when training data is small. 

\begin{figure}
\begin{center}
   \includegraphics[width=0.88\linewidth]{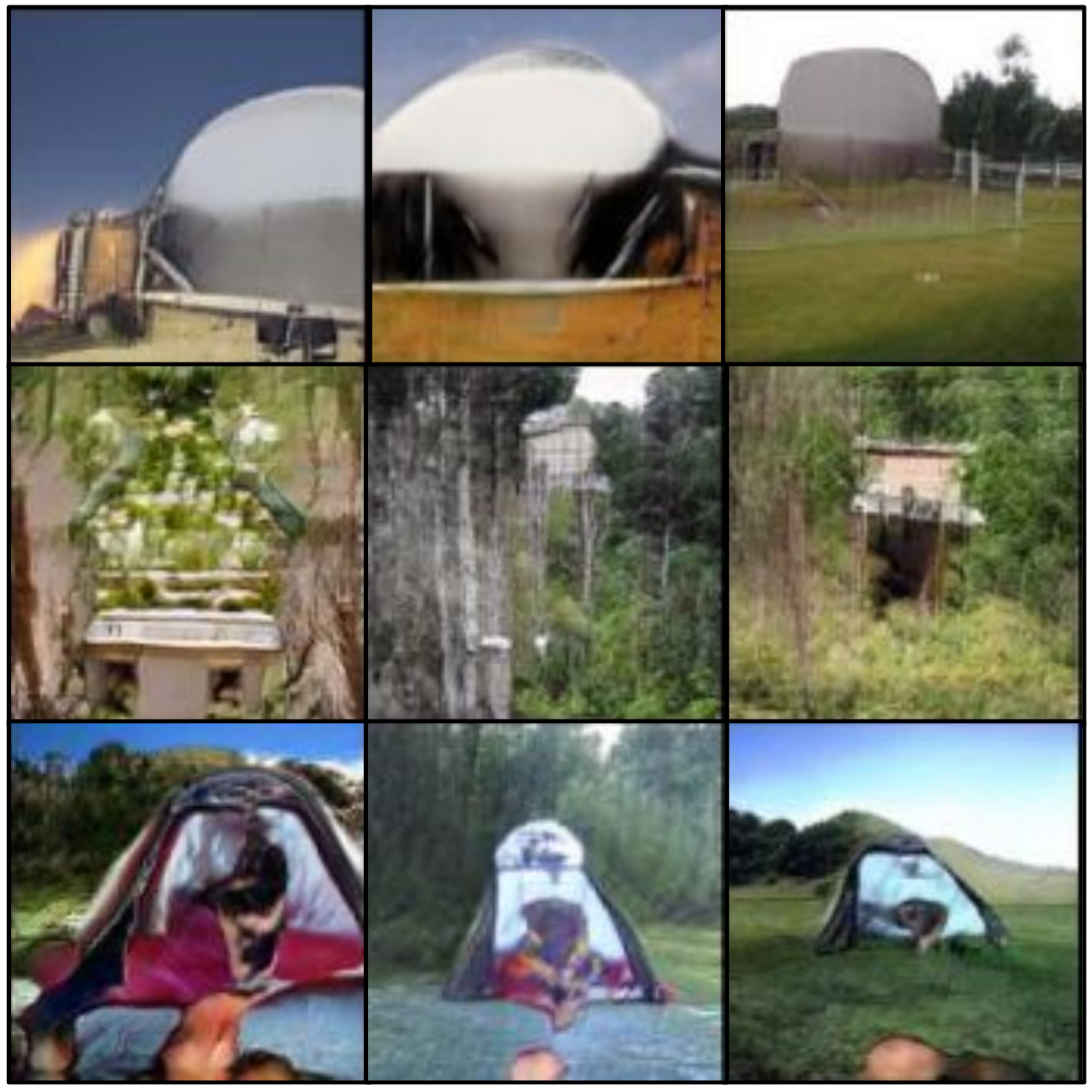}
\end{center}
   \caption{Top 3 contributing classes (planetarium, Bird House, Mountain Tent) from the pre-trained network toward the target class ``Arch" in places365 dataset~\cite{zhou2014places}. The classes are selected based on the learned similarity scores of the first layer. Each row depicts one class, and the images are generated from the network pre-trained on ImageNet.}
\label{fig:3arch}
\end{figure}
\begin{figure}
\begin{center}
   \includegraphics[width=0.88\linewidth]{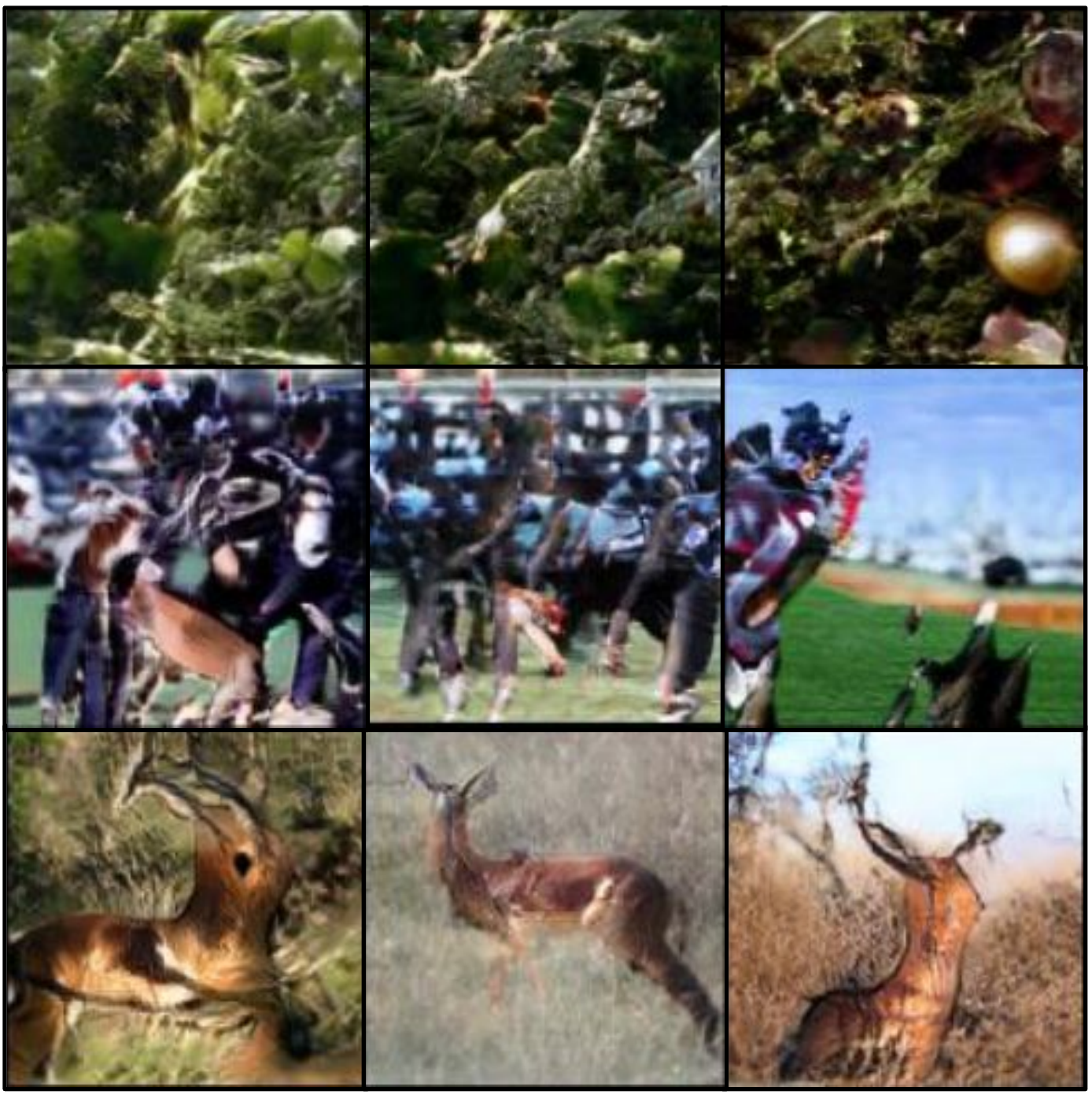}
\end{center}
   \caption{Top 3 contributing classes (Buckeye, Football Helmet, Impala) from the pre-trained network toward the target class ``Deer" in Animal Face dataset. The classes are selected based on the learned similarity scores of the first layer. Each row depicts one class, and the images are generated from the network pre-trained on ImageNet.}
\label{fig:3deer}
\end{figure}
\section{Discussion on Class Similarities}
As explained in the main paper, our method proposes knowledge transfer from previous classes by learning similarity scores over previous BN parameters and combing the BN parameters using those scores to construct the BN parameters of the new classes. Previous classes can contribute to a target class in terms of semantics, shape, texture, or color in a hierarchical manner from bottom to top layers. As an example, Fig. \ref{fig:3arch} shows the top 3 ImageNet classes of the pre-trained network (planetarium, Bird House, Mountain Tent) contributing to the target class ``Arch" in Places365, based on the similarity weights learned for the first layer (the first layer is generally more interpretable in terms of class similarities, since it is responsible for determining the general structure of the output images, as shown in Fig. 9 in the main paper). As it can be seen, these classes contain visual features close to arch structures that can meaningfully be used to generate images from the target class. Fig. \ref{fig:3deer} shows another example on Animal Face dataset~\cite{Zhangzhang2011animal} by visualizing the top 3 classes (Buckeye, Football Helmet, Impala) contributing to the target class ``Deer". In this example, we can see that the third class ``Impala" is semantically very close to the target class. However, the contribution of the first two classes is not as clear as the previous example. These classes might be contributing to the background, or this might be due to the fact that the similarity scores are actually learned to combine the pseudo-classes, and this does not always guarantee semantic similarity to the initial pre-training classes.
\section{FID and Loss Curves}
\begin{figure}[h]
\begin{center}
    \includegraphics[width=\linewidth]{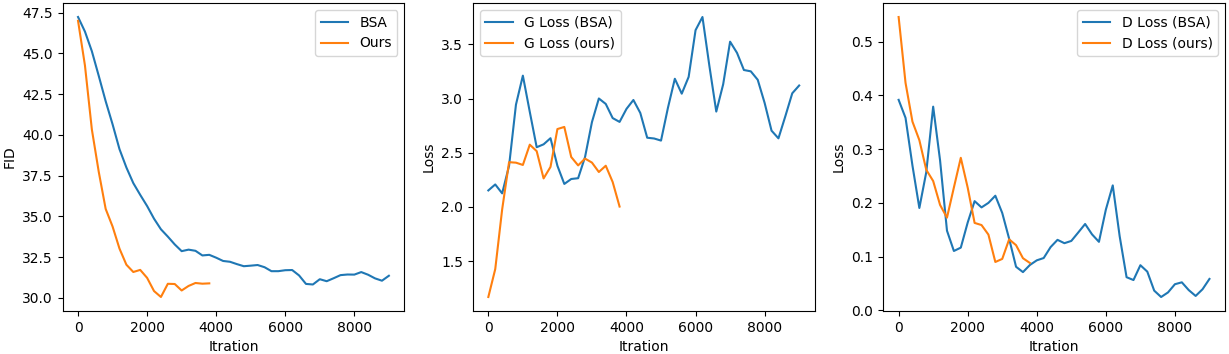}
\end{center}
\vspace{-0.4cm}
   \caption{The FID curve (left), the G loss (middle), and the D loss (right) for our method and BSA on CIFAR(10/600).}
\label{fig:long}
\label{fig:curve}
\vspace{-0.4cm}
\end{figure}

As an example of how the losses and the FID scores evolve during the training, the curves for on of the CIFAR100 experiments (Exp. 10/600) have been provided in Fig.~\ref{fig:curve}. The convergence speed-up is clearly depicted in the FID curve, whereas the same is difficult to be derived from the loss plots (due to the adversarial training).
\section{Single-class Target}\label{sec:single}
Although the main focus of our work is multi-class to multi-class knowledge transfer using knowledge propagation and knowledge sharing, the proposed method is not only limited to the multi-class target. As an example, the results of knowledge transfer to the single class ``Arch" in Places365 are provided in Table \ref{table:single}.
\begin{table}[h]
\begin{center}
\scalebox{0.9}{
\begin{tabular}{|l|c|c|}
\hline
Method & FID & Iterations\\
\hline\hline
BSA & 104 & 4300 \\
Ours & \textbf{78}  & \textbf{500} \\
\hline
\end{tabular}
}
\end{center}
\vspace{-0.4cm}
\caption{FID scores and number of iterations for knowledge transfer from ImageNet to the single target class ``Arch" in Places365.}
\label{table:single}
\end{table}
\section{Additional Visual Results}
\begin{figure*}
\begin{center}
   \includegraphics[width=\linewidth]{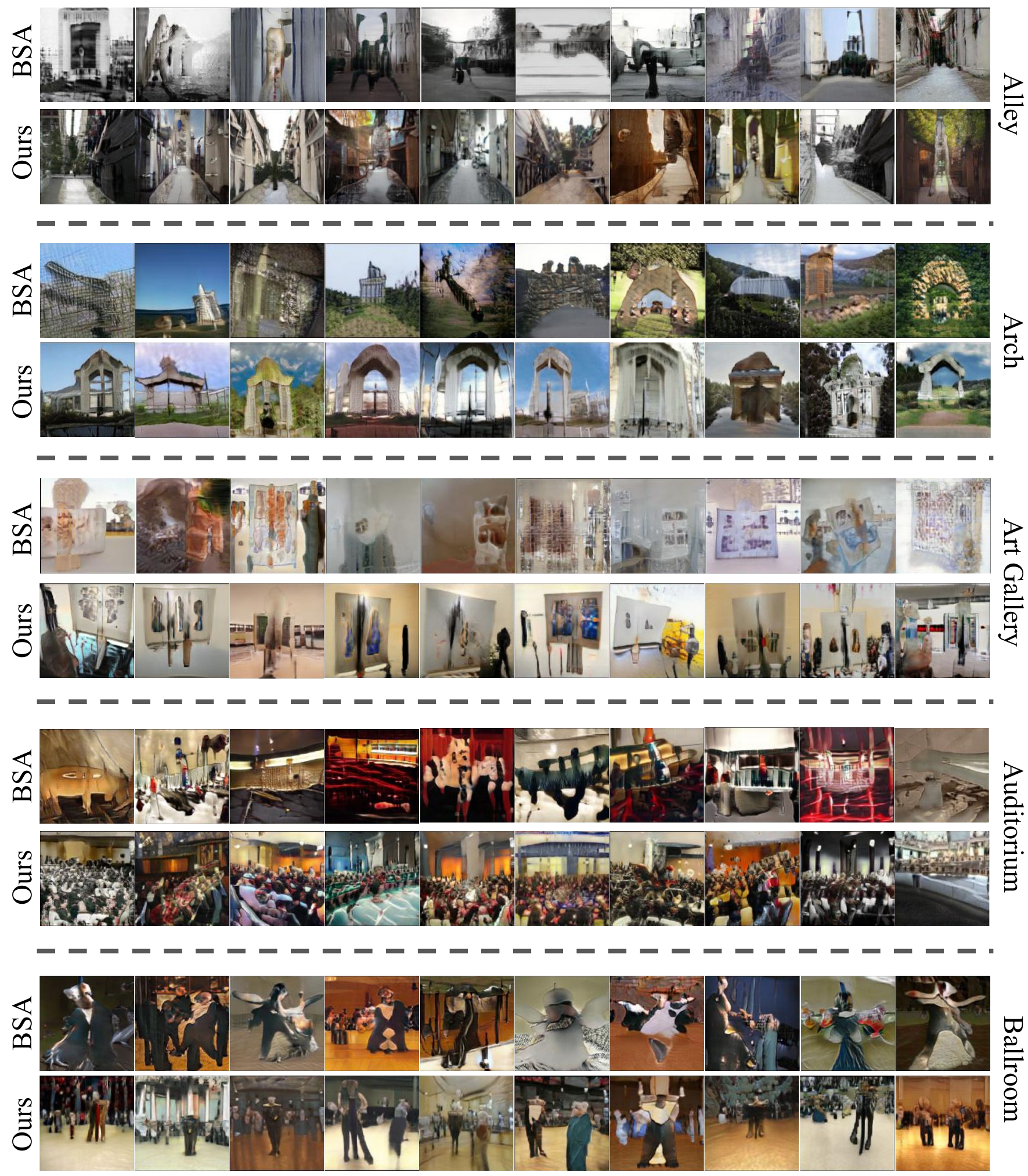}
\end{center}
\vspace{-0.4cm}
   \caption{Visual comparison between the images obtained from BSA (no knowledge transfer across classes) and our method on 5 classes of Places365.}
\label{fig:places_sup}
\end{figure*}

\begin{figure*}
\begin{center}
   \includegraphics[width=0.9\linewidth]{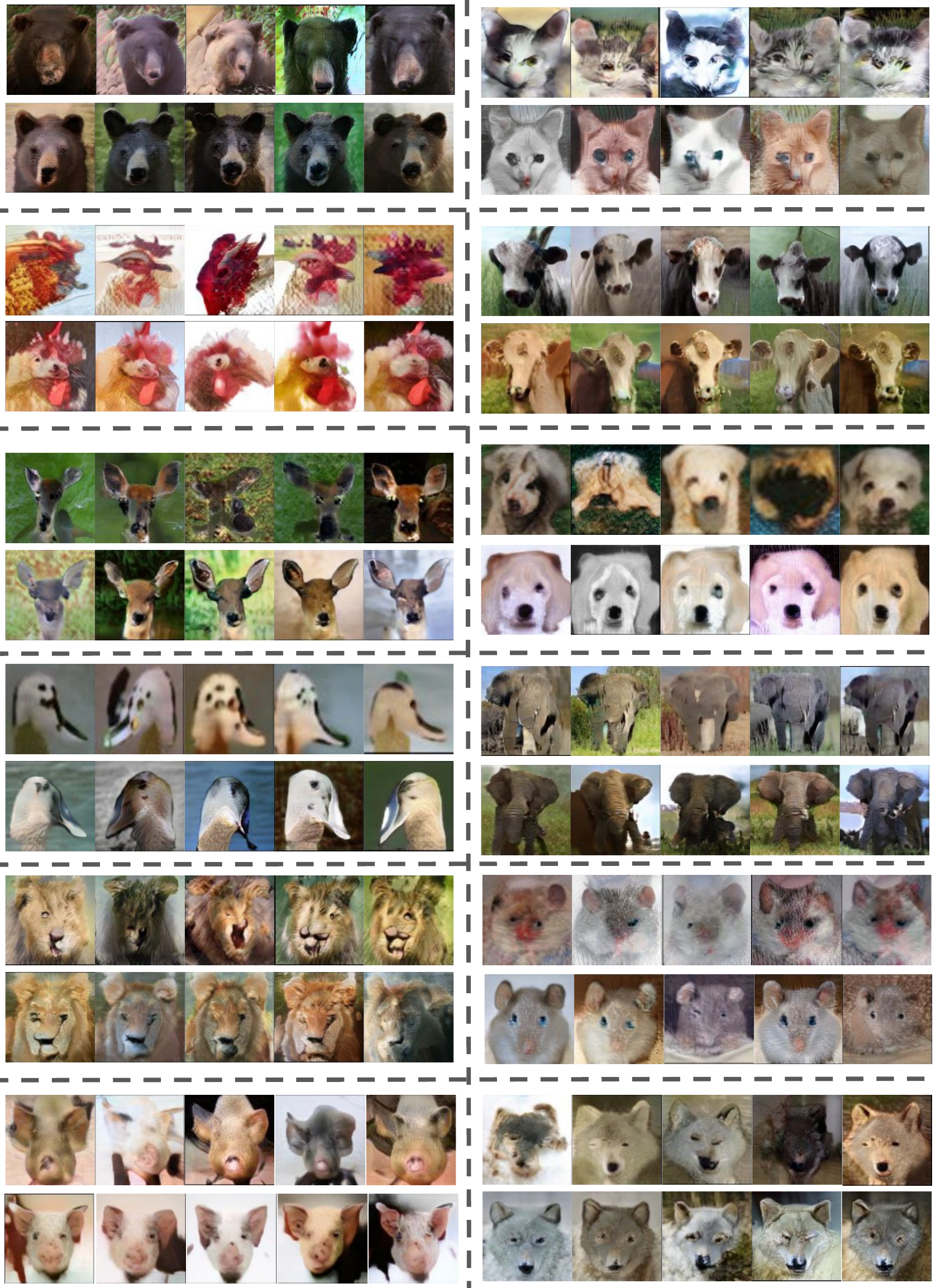}
\end{center}
\vspace{-0.4cm}
   \caption{Visual comparison between the images obtained from BSA (no knowledge transfer across classes) and our method on some of the classes of Animal Face (for each class, the first row is from BSA, and the second row from our method).}
\label{fig:animal_sup}
\end{figure*}

\begin{figure*}
\begin{center}
   \scalebox{0.45}{\includegraphics[width=\linewidth]{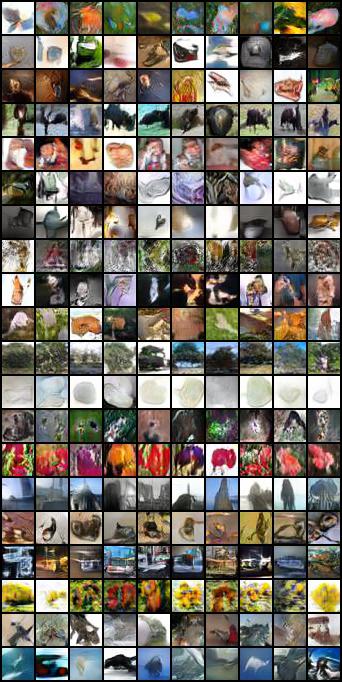}}
   \scalebox{0.45}{\includegraphics[width=\linewidth]{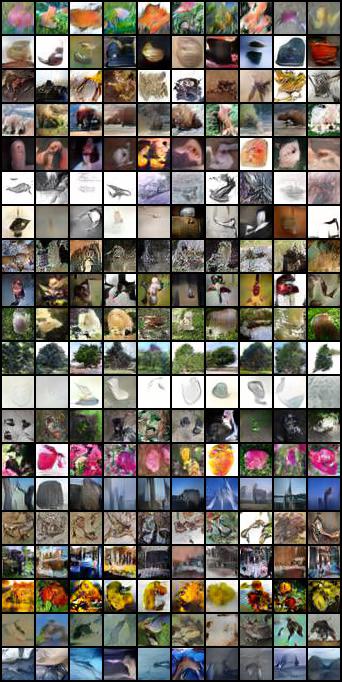}}
\end{center}
\vspace{-0.4cm}
   \caption{Visual comparison between the images obtained from BSA (left) and our method (right) for transferring from 80 classes of CIFAR100 to 20 classes each containing 300 samples.}
\label{fig:cifar20_300_sup}
\end{figure*}

\begin{figure*}
\begin{center}
   \scalebox{0.55}{\includegraphics[width=0.9\linewidth]{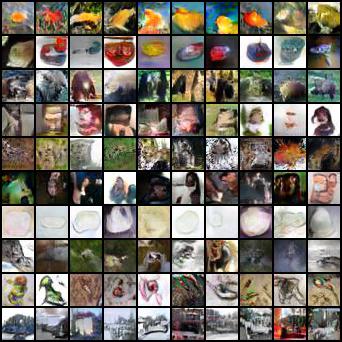}}
   \scalebox{0.55}{\includegraphics[width=0.9\linewidth]{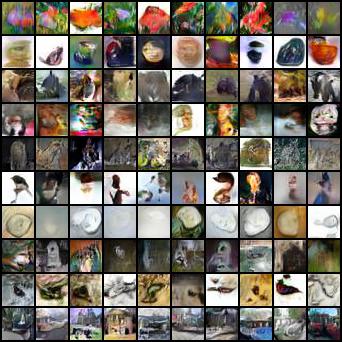}}
\end{center}
\vspace{-0.4cm}
   \caption{Visual comparison between the images obtained from BSA (left) and our method (right) for transferring from 80 classes of CIFAR100 to 10 classes each containing 300 samples.}
\label{fig:cifar10_300_sup}
\end{figure*}

Fig. \ref{fig:places_sup} and Fig. \ref{fig:animal_sup} show additional visual result obtained from BSA (no knowledge propagation across classes) and our method on ImageNet setup. Regarding CIFAR setup, we visualize the results of the experiments 20/300 and 10/300 for BSA and our method in Fig. \ref{fig:cifar20_300_sup} and Fig. \ref{fig:cifar10_300_sup}, as examples of CIFAR experiments.




\end{document}